\documentclass[10pt,twocolumn,letterpaper]{article}
\usepackage[pagenumbers]{cvpr}
\usepackage[T1]{fontenc}
\usepackage{amsmath,amssymb,booktabs,graphicx,tabularx,array,xcolor,listings}
\usepackage{microtype}
\usepackage{placeins}
\definecolor{cvprblue}{rgb}{0.21,0.49,0.74}
\usepackage[pagebackref,breaklinks,colorlinks,allcolors=cvprblue]{hyperref}
\hypersetup{pdftitle={SofaSpace: A Benchmark for Executable Visual Spatial Reasoning in Continuous Motion and Environment Construction},pdfauthor={Ze Rong, Kun Shi, Zhedong Zheng}}
\def\confName{CVPR}
\def\confYear{2027}

\title{SofaSpace: A Benchmark for Executable Visual Spatial Reasoning in\\Continuous Motion and Environment Construction}

\author{Ze Rong\qquad Kun Shi\qquad Zhedong Zheng\textsuperscript{*}\\
Faculty of Information Science and Computing, University of Macau\\
Macau, China\\
{\small \textsuperscript{*}Corresponding author}}

\newcommand{\figonemotivation}{%
  \begin{minipage}{\textwidth}
    \centering
    \includegraphics[width=0.77\textwidth]
      {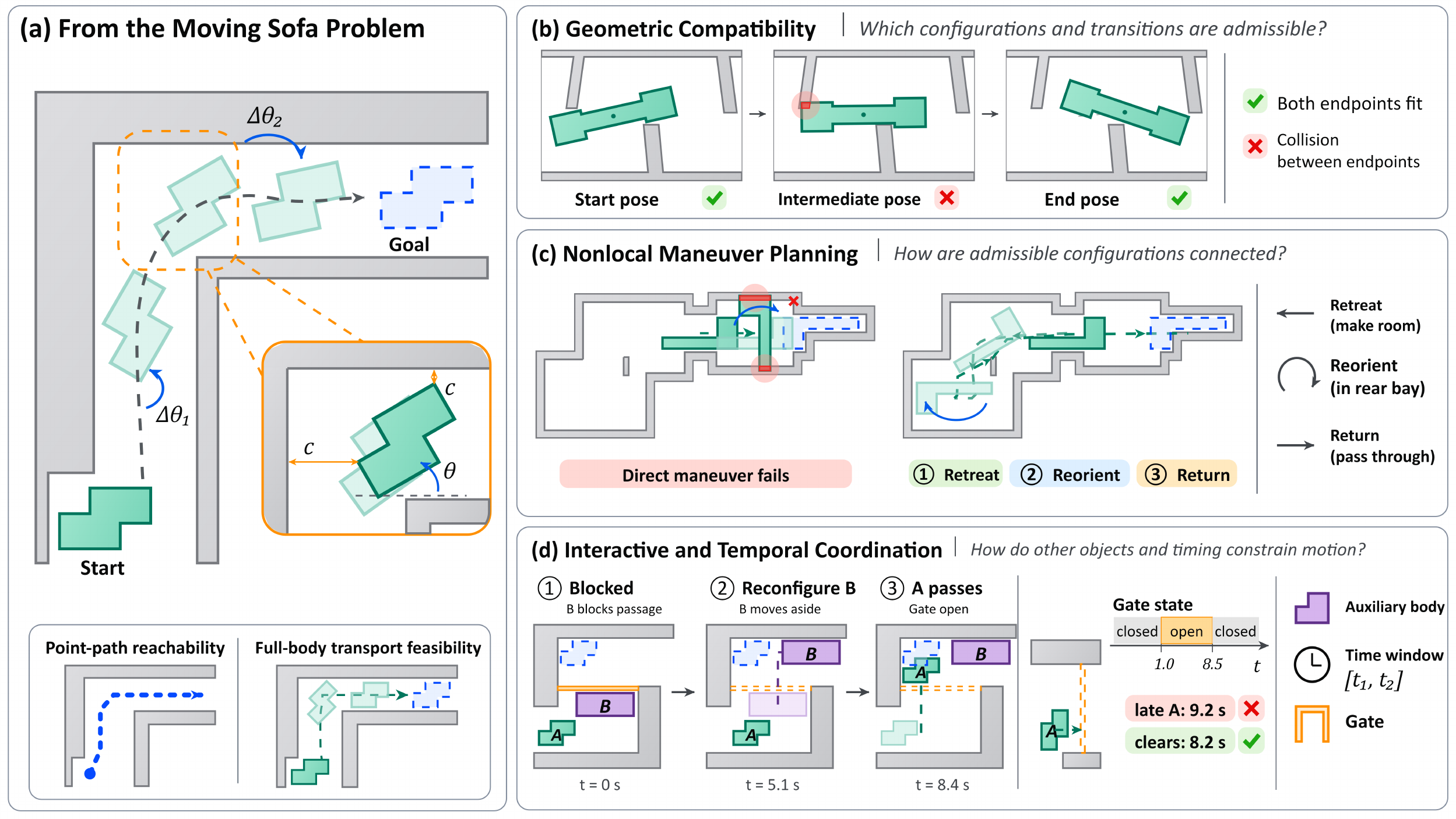}
\vspace{-.1in}
    \captionof{figure}{\textbf{What makes motion possible?}
    (a) Continuous rigid-body transport needs feasibility throughout.
(b) Geometric compatibility: admissible configurations and transitions.
(c) Nonlocal maneuver planning: reorientation, route choice or temporary retreat into a complete motion.
(d) Interactive and temporal coordination: dependence on other movable bodies or time-varying constraints.\\
    "We must perceive in order to move, but we must also move in order to perceive." - James J. Gibson}
    \label{fig:motivation}
  \end{minipage}%
}

\makeatletter
\g@addto@macro\@maketitle{%
  \vspace*{-6pt}%
  \figonemotivation
  \vspace*{3pt}%
}
\makeatother

\begin{document}

\maketitle
\begin{abstract}
\itshape
Planar rigid-body transport is to move a body from one pose to another through space that may barely admit the motion.
The moving sofa problem asks what makes such motion possible.
It is challenging in the nonlocality of the constraint, because a valid pose need not belong to a valid path.
Existing benchmarks concentrate on spatial relations, geometric transformations and visual planning, but underestimate continuous full-body motion, which they score at isolated configurations.
In this work, we argue that spatial intelligence should be measured on executable motion, not on the configurations a model names.
Specifically, we introduce a benchmark, called SofaSpace, to pose visual spatial reasoning as planar rigid-body transport, organized by three challenge classes.
Geometric compatibility concerns which configurations and transitions are admissible.
Nonlocal maneuver planning concerns how admissible configurations form a complete motion.
Interactive and temporal coordination captures dependencies on other movable bodies and time-varying constraints.
Verification executes the proposed motion, so a correct endpoint or an individually valid pose does not pass.
SofaSpace contains 3,000 scenes, and a separate 534-task SofaSpace-Hard set probes the capability frontier of current models.
The loss from configuration to motion is internal to a single plan: Claude reaches 97.2\% endpoint agreement but 52.8\% continuous validity, DeepSeek drops from 100.0\% to 30.6\%, and Kimi from 97.2\% to 29.4\%, while the strongest configuration retains 98.3\% full validity.
On SofaSpace-Hard the best model resolves 5 of 19 feasibility judgments,
while Grok resolves 1 and the remaining configurations resolve none.
Besides, geometric feasibility does not imply functional compliance: 181 of 295 feasible layouts satisfy all evaluated functional requirements for the strongest model, and 1 of 172 for Kimi.
Because verification needs only a simulator and a collision check, the protocol extends from planar bodies to articulated and three-dimensional transport.
\end{abstract}

\section{Introduction}
\label{sec:intro}

The classical moving sofa problem seeks the largest planar shape that can traverse a right-angled corner~\cite{romik2016movingsofa}.
The shape of the object remains compatible with the surrounding space throughout translation and rotation.
The closely related piano movers problem formalizes this for a fixed rigid body moving between prescribed configurations among obstacles~\cite{schwartz1983pianomovers}.
These problems motivate a basic question for visual spatial intelligence:
\emph{what makes motion possible?}

Spatial intelligence extends beyond representing where objects are to reasoning about what actions the surrounding space permits and how those possibilities evolve as an object moves.
Human spatial cognition connects such reasoning to affordance perception, mental simulation, and prospective action planning~\cite{warren1987apertures,hegarty2004mechanical,rosenbaum1992planning}.
Rigid-body transport integrates these capabilities by requiring reasoning over a sequence of geometrically constrained configurations.
Though recent benchmarks cover planning, spatial transformation, assembly, and continuous geometric reasoning~\cite{wu2024vsp,wang2026spatialvizbench,jiang2025marble,zong2026tangramsr}, continuous full-body transport rarely serves as the primary target of executable verification.
This gap raises three questions: Can current VLMs reason about the constraints underlying feasible rigid-body transport?
How does performance vary under nonlocal, interactive, or temporal demands?
Can models construct environments that realize prescribed transport behavior in addition to planning within a given space?

To study these questions, we introduce \textbf{SofaSpace}, a benchmark for visual spatial reasoning grounded in planar rigid-body transport.
The planar setting renders geometry and motion explicitly verifiable while retaining requirements such as reorientation, detours, and interaction.
We organize transport difficulty into three complementary challenges.
\textbf{Geometric compatibility} concerns which configurations and transitions are admissible, \textbf{nonlocal maneuver planning} concerns how admissible configurations form a complete motion, and \textbf{interactive and temporal coordination} captures dependencies on other movable bodies and time-varying constraints.
Seven composable scene mechanisms instantiate these challenges and their combinations.

These challenge classes characterize what makes transport difficult, while a complementary task axis specifies how models reason over the same constraints.
SofaSpace probes the relation between geometry and motion in both directions.
Given an environment, models reconstruct its geometry, judge motion validity or feasibility, or construct a complete motion.
Given prescribed transport requirements, models instead construct an environment and, jointly, a valid motion within it.
The 3,000-scene \textbf{SofaSpace} supports aligned evaluation across reconstruction, motion planning, layout generation, and joint layout and motion generation.
A separate \textbf{SofaSpace-Hard} set provides more demanding planning and judgment problems for probing the capability frontier of current models.
Motion outputs are evaluated through continuous full-body verification, and generated layouts by geometric and functional constraint checking.

We evaluate diverse vision-language and reasoning models and observe distinct failure patterns.
Motion planning failures arise both in rigid-body configurations and in the continuous motion between them, showing that configuration-level correctness does not necessarily imply executable transport.
Reconstruction reveals coarse structural agreement fails to guarantee motion-semantic fidelity:
reconstructed environments can preserve high-level structure while changing which motions are admitted or rejected.
Under our strict criterion, geometry-correct reconstructions preserve all decided motion probes.
Environment construction further separates geometric feasibility, functional compliance, and self-execution, as a feasible layout may fail the prescribed transport requirements and a functionally valid layout may still be paired with an invalid submitted motion.
Together, these results motivate evaluating spatial competence through how models represent, preserve, and construct the constraints that make motion possible. Our contributions are as follows:
\begin{itemize}
    \item Different from benchmarks that score isolated configurations, \textbf{SofaSpace} is a visual spatial reasoning benchmark that evaluates the complete motion of an extended rigid body, grouping instances into 3 challenge classes instantiated by 7 composable scene mechanisms.

    \item Motivated by the gap between a legal endpoint and a legal motion, every answer is verified by interpolating between the submitted poses and checking full-body clearance and timing over each segment.
    Four task views share that criterion, reconstructing, judging, planning in and constructing the same scenes, over a 3,000-scene Core and a 534-task \textbf{SofaSpace-Hard} set.

    \item We observe three gaps that point the same way: endpoint agreement above 97\% yields only 29.4\% to 52.8\% executable motion, 97.33\% structural agreement still rejects 6.26\% of valid motions, and the best configuration resolves 5 of 19 feasibility judgments while Grok resolves 1 and the remaining configurations resolve none.
\end{itemize}

\begin{table*}[t]
\centering
\caption{\textbf{Comparison with representative spatial reasoning, planning, and construction works.}
We summarize each work using a shared task taxonomy and output space.
Multi-step denotes reasoning over multiple dependent steps; Temporal denotes explicit temporal reasoning;
Functional denotes reasoning about functional requirements or affordances;
Continuous denotes continuous geometric configurations or motion;
and Construction indicates that environment geometry is generated as an output.}
\label{tab:related_comparison}
\vspace{-.1in}
\footnotesize
\setlength{\tabcolsep}{2.2pt}
\renewcommand{\arraystretch}{1.12}

\begin{tabular*}{\textwidth}{
@{\extracolsep{\fill}}
l|
l
l
c
c
c
c
c
}
\toprule
Work &
Task Type &
Output &
Multi-step &
Temporal &
Functional &
Continuous &
Construction \\
\midrule

VSP~\cite{wu2024vsp}
& Reasoning / Planning
& Actions / Answers
& $\checkmark$
& $\checkmark$
& --
& --
& -- \\

MARBLE~\cite{jiang2025marble}
& Reasoning / Planning
& Plans / Answers
& $\checkmark$
& --
& --
& --
& -- \\

Flat-Pack~\cite{chetan2026flatpack}
& Reasoning
& Answers
& $\checkmark$
& $\checkmark$
& --
& --
& -- \\

SFI-Bench~\cite{zhang2026sfibench}
& Reasoning
& Answers
& $\checkmark$
& $\checkmark$
& $\checkmark$
& --
& -- \\

TangramSR~\cite{zong2026tangramsr}
& Reasoning / Generation
& Configurations
& --
& --
& --
& $\checkmark$
& -- \\

PhyScene~\cite{yang2024physcene}
& Generation
& Layouts
& --
& --
& $\checkmark$
& $\checkmark$
& $\checkmark$ \\

\midrule

\textbf{SofaSpace}
& \textbf{Reasoning / Planning / Generation}
& \textbf{Answers / Plans / Layouts}
& $\checkmark$
& $\checkmark$
& $\checkmark$
& $\checkmark$
& $\checkmark$ \\

\bottomrule
\end{tabular*}
\vspace{-.1in}
\end{table*}

\section{Related Work}
\label{sec:related}

\noindent\textbf{Visual Spatial Reasoning Benchmarks.}
Multiple benchmarks extend spatial reasoning beyond static relations~\cite{chen2024spatialvlm,cheng2024spatialrgpt,wang2024spatialeval,wang2025spatial457,ma2025_3dsrbench}. MMSI-Bench and VSI-Bench study spatial understanding across multiple visual observations and extended environments~\cite{yang2026mmsibench,yang2025vsibench,zhu2025cospace}, while SpatialViz-Bench evaluates transformations such as mental rotation and folding~\cite{wang2026spatialvizbench}. RoboSpatial, SFI-Bench, Flat-Pack Bench, and GeoText-1652 connect spatial reasoning with reference frames, placement compatibility, and temporal ordering~\cite{chu2024towards,song2025robospatial,zhang2026sfibench,chetan2026flatpack}. SofaSpace instead makes complete rigid-body motion the object of executable evaluation (Table~\ref{tab:related_comparison}).

\noindent\textbf{Geometric Motion Planning and Rigid-Body Transport.} Rigid-body transport has a classical foundation in computational geometry and robotics. The moving sofa and piano movers problems characterize how object geometry, surrounding space, and prescribed configurations constrain continuous motion~\cite{romik2016movingsofa,schwartz1983pianomovers}, and motion planning casts transport as a collision-free path through configuration space~\cite{lavalle2006planning}. Recent benchmarks approach related problems through navigation, interactive planning, constrained path finding, assembly, and continuous geometric configuration~\cite{wu2024vsp,mayer2025ivispar,kaesberg2025sparc,jiang2025marble,zong2026tangramsr}. SofaSpace instead requires geometric and temporal validity along the whole connecting motion.

\noindent\textbf{Spatial Generation and Verifiable Construction.} Spatial generation constructs environments under semantic, geometric, or physical constraints. LayoutGPT, Holodeck, and PhyScene generate structured or interactable environments from prescribed conditions~\cite{feng2023layoutgpt,yang2024holodeck,yang2024physcene,sun2025layoutvlm}, while InfiniBench generates controllable scenes and camera trajectories for spatial reasoning evaluation~\cite{wang2026infinibench}. Executable verification also appears in problem construction and learning, as in Absolute Zero, GeoX, and Ouroboros-Spatial~\cite{zhao2025absolutezero,ahn2026geox,zhao2026ouroborosspatial}. SofaSpace treats construction as the inverse of planning: the generated environment should admit a verified complete motion that meets the fixed transport requirements.

\begin{figure*}[t]
    \centering
    \vspace{-.1in}
    \includegraphics[width=0.95\textwidth]{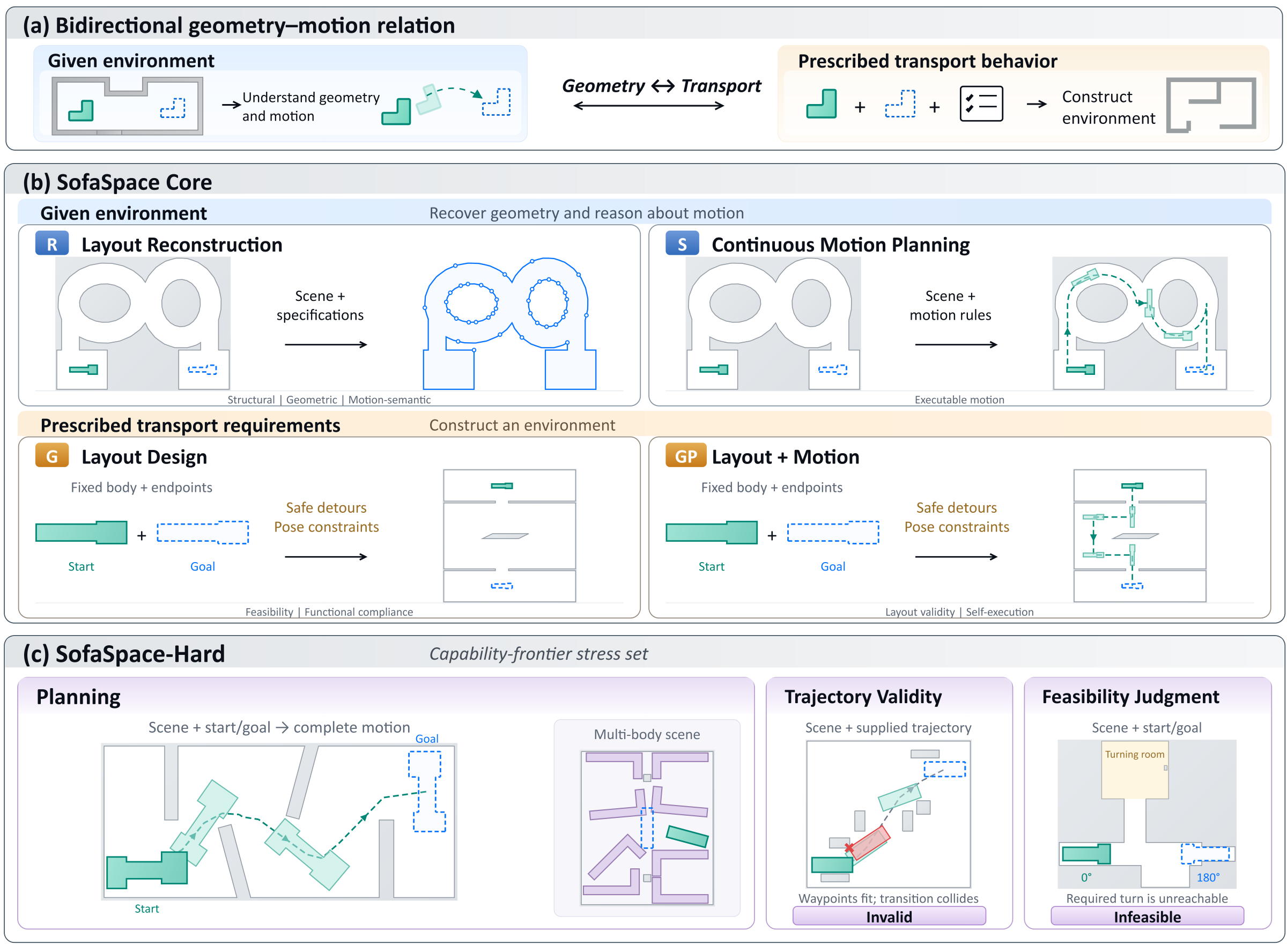}
    \vspace{-.2in}
    \caption{\textbf{SofaSpace task overview.}
    (a) SofaSpace studies the bidirectional relation between environment geometry and feasible rigid-body transport.
    (b) The SofaSpace contains four aligned task views: R and S reason from a given environment, whereas G and GP construct environments from prescribed transport requirements.
    (c) SofaSpace-Hard contains planning, trajectory-validity, and feasibility problems for probing the capability frontier of current models.}
    \label{fig:task_overview}
    \vspace{-.1in}
\end{figure*}

\section{The SofaSpace Benchmark}
\label{sec:benchmark}

SofaSpace contains 3,000 scenes with four aligned task views over the geometry--motion relation, while the separate \textbf{SofaSpace-Hard} set provides more demanding planning and validity/feasibility problems.

\subsection{Benchmark Tasks}
\label{sec:task_views}

The four SofaSpace task views share scene identities but expose different public information and outputs, as summarized in Fig.~\ref{fig:task_overview}.
S receives a complete scene image and task instructions and returns a motion plan, while R additionally receives fixed-body and endpoint context and reconstructs the environment geometry.
G and GP receive fixed bodies, endpoints, and functional requirements while the target layout remains hidden.
Reference layouts, author motions, and validation certificates are private.

\noindent\textbf{Solve and reconstruct.}
S requires a complete motion satisfying the geometric, temporal, and multi-body constraints; SofaSpace-Hard additionally includes trajectory-validity and feasibility judgments.
R reconstructs free-space and gate geometry, which we evaluate for structural, geometric, and motion-semantic fidelity.

\noindent\textbf{Generate and plan.}
G constructs an environment while keeping the supplied bodies and endpoint configurations fixed; a valid layout should be geometrically legal, satisfy the prescribed functional conditions, and admit a verified complete motion.
Because alternative layouts are admissible, correctness depends on supported transport behavior rather than similarity to a single reference layout.
GP additionally requires a valid submitted motion; detailed schemas and task-specific acceptance criteria are provided in Appendix A of the supplementary material.

\begin{figure*}[t]
    \centering
    \vspace{-.1in}
    \includegraphics[width=\textwidth]{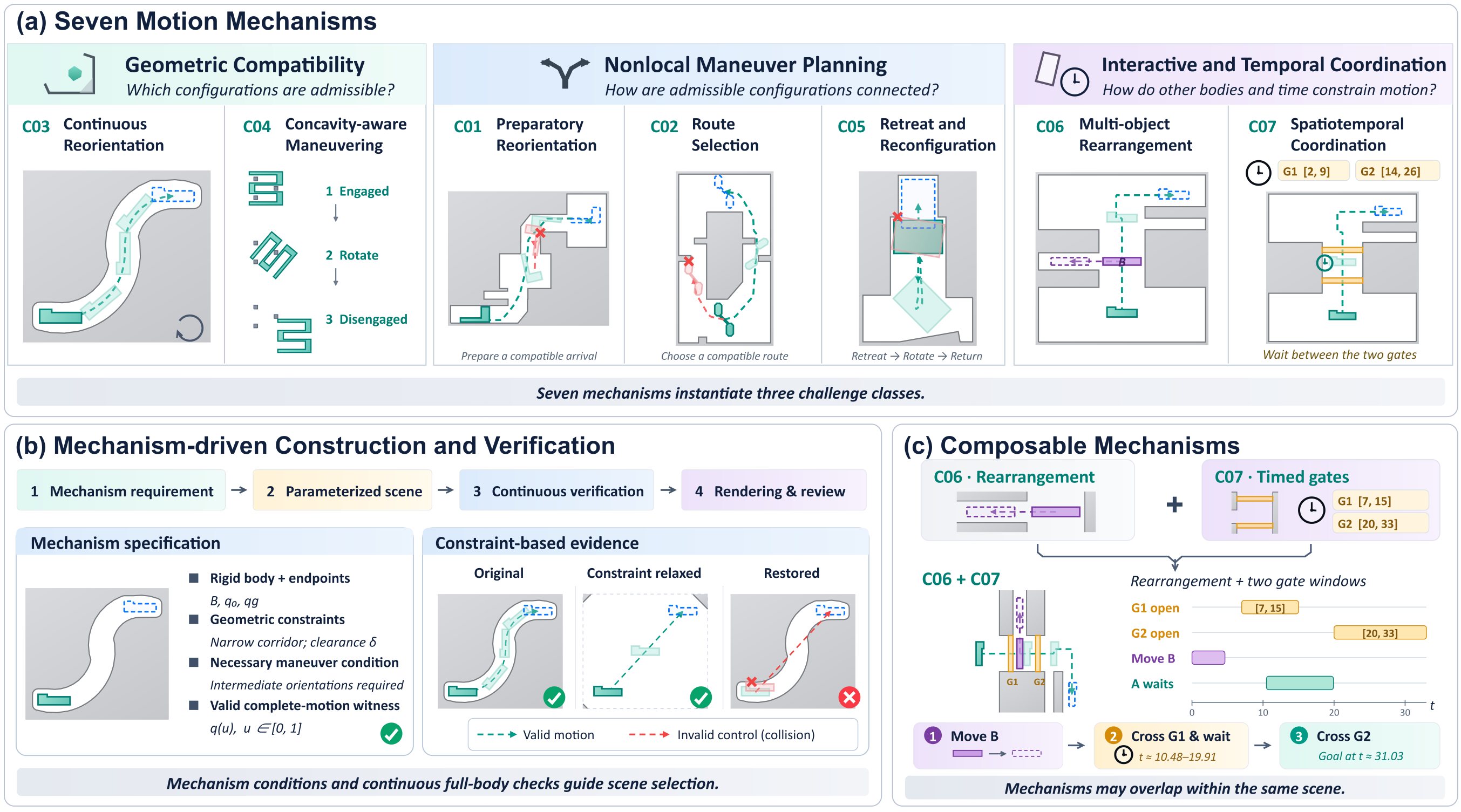}
    \vspace{-.25in}
    \caption{\textbf{Constraint mechanisms in SofaSpace.}
    (a) 7 composable motion mechanisms instantiate the 3 challenge classes.
    (b) Mechanism requirements guide parameterized scene construction and verification through constraint-specific geometric and motion evidence.
    (c) Mechanisms may compose within a scene; the example combines multi-object rearrangement (C06) with spatiotemporal coordination (C07).}
    \label{fig:mechanisms}
    \vspace{-.1in}
\end{figure*}

\subsection{Executable Motion and Verification}
\label{sec:motion_specification}

We formulate transport in configuration space~\cite{lavalle2006planning}.
An instance specifies planar free space, rigid bodies $B_1,\ldots,B_m$,
initial and goal poses, and motion limits.
At time $t$, body $i$ has a planar rigid-body pose represented by
$(x_i(t),y_i(t),\theta_i(t))$ in $\mathrm{SE}(2)$ and occupies $
P_i(t)=R_{\theta_i(t)}B_i+(x_i(t),y_i(t))^\top.$

Let $F(t)$ denote traversable free space, accounting for time-varying
obstacles such as closed gates.
Suppressing the common time argument for compactness, a valid motion at
clearance $\delta>0$ satisfies
\begin{equation}
d(P_i,F^c)\geq\delta,\qquad
d(P_i,P_j)\geq\delta\quad(i\neq j).
\label{eq:full_body_validity}
\end{equation}
Here, $F^c$ denotes the complement of free space in the plane,
and $d$ is the Euclidean distance between sets.
The conditions hold throughout every motion segment and ensure
full-body containment and clearance.

Positions and explicitly unwrapped orientations are continuously
interpolated between submitted poses.
Timed and multi-body tasks additionally account for waiting, gate
schedules, and coordinated body motion.
All segments are conservatively checked for full-body occupancy
and the applicable temporal and pairwise constraints.
Numerical tolerances and output contracts are detailed in
Appendix A.

\subsection{Challenge Classes and Mechanisms}
\label{sec:constraint_mechanisms}

SofaSpace organizes transport difficulty into three complementary challenge classes instantiated by seven composable mechanisms, summarized in Fig.~\ref{fig:mechanisms}.
\textbf{Geometric compatibility} concerns whether body shape, orientation, and free space admit the required configurations and transitions (C03--C04).
\textbf{Nonlocal maneuver planning} requires organizing admissible configurations into a complete motion through reorientation, route choice, or temporary retreat (C01, C02, and C05), while \textbf{interactive and temporal coordination} captures dependencies on movable bodies or time-varying constraints (C06--C07).
Mechanisms may overlap and should be interpreted as scene-construction requirements rather than mutually exclusive or universally calibrated difficulty levels.
Mechanism-specific requirements are checked using geometric and motion evidence; detailed certificates and constraint-release controls are provided in Appendix B.

\subsection{Construction, Quality Control, and Splits}
\label{sec:benchmark_composition}

SofaSpace scenes are constructed from mechanism-driven parent families and parameterized variants, with admission based on geometric legality, verified motion feasibility, mechanism-specific evidence, and visual review.
Because certificates are interface-specific, task-view eligibility may differ across scenes; we report the resulting composition in Table~\ref{tab:counts}.
Registered parent families are disjoint across the SofaSpace splits, with derivatives grouped together.
The 300-scene test set contains 180 unseen-family scenes with familiar body constructions, 60 scenes with held-out gate-dependency structures, and 60 scenes jointly holding out the paired-cavity body construction and its associated family.
SofaSpace-Hard separately contains 534 tasks: 473 planning problems, 42 trajectory-validity judgments, and 19 feasibility judgments, evaluated with its task-specific interfaces and acceptance criteria; detailed provenance and rules are provided in Appendix E.

\begin{table}[t]
    \centering
    \caption{\textbf{Composition and view eligibility.}
    S denotes Continuous Motion Planning, R denotes Layout Reconstruction,
G denotes Layout Design, while GP represents Layout and Motion.}
    \label{tab:counts}
    \vspace{-.1in}
    \small
    \renewcommand{\arraystretch}{1.08}
    \setlength{\tabcolsep}{0pt}

    \begin{tabularx}{\columnwidth}{
        @{}
        >{\raggedright\arraybackslash}p{0.24\columnwidth}
        >{\centering\arraybackslash}p{0.17\columnwidth}
        >{\centering\arraybackslash}p{0.13\columnwidth}
        >{\centering\arraybackslash}p{0.13\columnwidth}
        >{\centering\arraybackslash}X
        @{}
    }
        \toprule
        Split & Instances & S & R & G / GP \\
        \midrule
        Train      & 2,600 & 2,600 & 2,600 & 2,564 / 2,599 \\
        Dev        & 100   & 100   & 100   & 100 / 100 \\
        Test       & 300   & 300   & 300   & 300 / 300 \\
        \midrule
        total & 3,000 & 3,000 & 3,000 & 2,964 / 2,999 \\
        Hard       & 534   & ---   & ---   & --- \\
        \bottomrule
    \end{tabularx}
    \vspace{-.1in}
\end{table}

\section{Experiment}
\label{sec:experiment}

\subsection{Experimental Setup}
\label{sec:exp_setup}

\noindent\textbf{Models and protocol.}
We evaluate seven contemporary API model configurations and the original Qwen3.5-9B checkpoint in a zero-shot setting:
GPT-6 Astra, Claude Fable 5.1, DeepSeek V4.1 Flash, Gemini 4 Argon, GLM-5.3-Flash, Grok 4.7, Kimi K3, and Qwen3.5-9B.
For each view, models receive only the corresponding public image, task prompt, and fixed context; no demonstrations, private reference layouts, author trajectories, or validation certificates are provided.
All eight configurations are evaluated on the complete 534-task SofaSpace-Hard set under the same Hard evaluation protocol.
For S, R, G, and GP, the primary metric is strict task success
under the view-specific acceptance criteria.
For R, strict success requires satisfying the frozen geometric
reconstruction contract; we separately analyze coarse structural
agreement (SA) and motion-semantic fidelity in Section~4.4.
Incomplete or length-terminated responses contribute no strict successes,
and all diagnostic analyses replay the frozen model outputs without
repairing submitted geometry or motion.

\noindent\textbf{Feasibility-search reliability.}
Because G requires establishing the existence of a complete motion without receiving a candidate trajectory, we audit the feasibility search on layouts with independently verified motion witnesses.
After withholding these witnesses, the frozen G search recovers 300/300 reference layouts and 925/942 (98.20\%) model-generated GP layouts.
Doubling both the time budget and internal search limits changes neither recovery rate nor any per-item decision, indicating that the reported G evaluation is stable with respect to the search budget used in our experiments. We request the highest reasoning setting used in our evaluation for each API configuration: Astra and Grok use xhigh reasoning, Claude uses adaptive thinking at maximum effort, DeepSeek and GLM-5.3-Flash use maximum thinking effort, Gemini uses its high-reasoning endpoint, and Kimi uses maximum effort. Qwen3.5-9B runs with thinking enabled. These provider-specific effort labels do not represent normalized compute budgets.

\begin{table}[t]
\centering
\caption{\textbf{Overall performance on SofaSpace.}
S denotes continuous motion planning, R represents layout reconstruction, G denotes layout design, GP denotes joint layout and motion generation, and Hard indicates the SofaSpace-Hard planning and judgment set. Strict success is reported for all SofaSpace views and SofaSpace-Hard. For R, strict success requires satisfying the frozen geometric reconstruction acceptance criteria. All values are in percentage; higher is better.}
\label{tab:main_results}
\scriptsize
\renewcommand{\arraystretch}{1.08}
\setlength{\tabcolsep}{1.2pt}

\begin{tabularx}{\columnwidth}{
@{}
>{\raggedright\arraybackslash}p{0.38\columnwidth}
>{\centering\arraybackslash}X
>{\centering\arraybackslash}X
>{\centering\arraybackslash}X
>{\centering\arraybackslash}X
>{\centering\arraybackslash}X
@{}
}
\toprule
Model & S & R Strict & G & GP & Hard \\
\midrule

GPT-6 Astra~\cite{openai2026gpt6astra}
& \textbf{99.00}
& \textbf{26.67}
& \textbf{60.33}
& \textbf{58.33}
& \textbf{90.64} \\

Claude Fable 5.1~\cite{anthropic2026fable51}
& 68.00
& 0.00
& 3.67
& 2.33
& 47.19 \\

DeepSeek V4.1 Flash~\cite{deepseek2026v41flash}
& 47.67
& 0.00
& 24.33
& 6.67
& 7.87 \\

Gemini 4 Argon~\cite{google2026gemini4argon}
& 12.00
& 9.33
& 46.00
& 28.33
& 14.04 \\

GLM-5.3-Flash~\cite{zai2026glm53flash}
& 31.67
& 0.00
& 10.00
& 8.33
& 5.81 \\

Grok 4.7~\cite{xai2026grok47}
& 49.33
& 0.00
& 51.00
& 39.33
& 7.30 \\

Kimi K3~\cite{moonshot2026kimik3}
& 47.67
& 0.00
& 0.33
& 0.00
& 5.43 \\

Qwen3.5-9B~\cite{qwen2026qwen35}
& 4.33
& 0.00
& 3.67
& 2.33
& 2.06 \\

\bottomrule
\end{tabularx}
\vspace{-.1in}
\end{table}

\begin{table*}[t]
\centering
\caption{\textbf{Mechanism-wise and compositional strict accuracy (\%).}
SofaSpace rows are mutually exclusive exact compositions over 300 test scenes; Hard atomic rows are non-exclusive memberships, while Hard composition rows partition all 534 tasks.
$N$ denotes group size.}
\label{tab:mechanism_composition}
\vspace{-.1in}
\footnotesize
\renewcommand{\arraystretch}{1.03}
\setlength{\tabcolsep}{0pt}

\begin{tabularx}{\textwidth}{
@{}
>{\raggedright\arraybackslash}p{0.135\textwidth}
>{\centering\arraybackslash}p{0.047\textwidth}
>{\centering\arraybackslash}X
>{\centering\arraybackslash}X
>{\centering\arraybackslash}X
>{\centering\arraybackslash}X
>{\centering\arraybackslash}X
>{\centering\arraybackslash}X
>{\centering\arraybackslash}X
>{\centering\arraybackslash}X
@{}
}
\toprule
Group & $N$ & Astra & Claude & DeepSeek & Gemini & GLM-5.3 & Grok & Kimi & Qwen 9B \\
\midrule

\multicolumn{10}{@{}l}{\textbf{SofaSpace test: exact mechanism composition}} \\
C03             & 60 & 100.0 & 80.0 & 53.3 & 5.0 & 43.3 & 53.3 & 50.0 & 21.7 \\
C04             & 60 & 96.7  & 10.0 & 0.0  & 1.7 & 0.0  & 0.0 & 0.0  & 0.0 \\
C06             & 30 & 100.0 & 96.7 & 83.3 & 86.7 & 33.3 & 93.3 & 86.7 & 0.0 \\
C07             & 30 & 100.0 & 93.3 & 83.3 & 0.0 & 63.3 & 83.3 & 90.0 & 0.0 \\
C01+C03         & 60 & 98.3  & 68.3 & 38.3 & 10.0 & 35.0 & 38.3 & 38.3 & 0.0 \\
C01+C06         & 30 & 100.0 & 83.3 & 73.3 & 0.0 & 26.7 & 83.3 & 86.7 & 0.0 \\
C01+C07         & 30 & 100.0 & 90.0 & 53.3 & 0.0 & 36.7 & 50.0 & 36.7 & 0.0 \\

\midrule
\multicolumn{10}{@{}l}{\textbf{SofaSpace-Hard: atomic mechanism membership}} \\
C01             & 242 & 93.4 & 44.6 & 2.5  & 10.3 & 1.2  & 2.9 & 1.7  & 0.4 \\
C02             & 77  & 93.5 & 22.1 & 2.6  & 2.6  & 1.3  & 1.3 & 0.0  & 0.0 \\
C03             & 151 & 94.0 & 53.0 & 15.9 & 17.2 & 13.9 & 14.6 & 11.9 & 5.3 \\
C04             & 83  & 85.5 & 28.9 & 0.0  & 4.8  & 0.0  & 0.0 & 0.0  & 0.0 \\
C05             & 203 & 94.6 & 40.9 & 1.5  & 8.4  & 1.5  & 1.5 & 0.0  & 0.0 \\
C06             & 136 & 91.9 & 53.7 & 6.6  & 18.4 & 2.9  & 5.1 & 4.4  & 0.7 \\
C07             & 54  & 92.6 & 63.0 & 11.1 & 14.8 & 7.4  & 18.5 & 11.1 & 3.7 \\

\midrule
\multicolumn{10}{@{}l}{\textbf{SofaSpace-Hard: mechanism composition}} \\
Single          & 183 & 81.4 & 56.3 & 18.6 & 24.0 & 14.2 & 16.9 & 13.1 & 5.5 \\
2-way           & 290 & 95.2 & 45.2 & 2.8  & 10.3 & 1.7  & 1.7 & 1.7  & 0.3 \\
3-way           & 61  & 96.7 & 29.5 & 0.0  & 1.6  & 0.0  & 4.9 & 0.0  & 0.0 \\

\bottomrule
\end{tabularx}
\vspace{-.1in}
\end{table*}

\subsection{Performance and Mechanism Composition}
\label{sec:main_results}

We report the overall results in Table~\ref{tab:main_results}.
The four SofaSpace views expose substantially different capability profiles.
Astra performs strongest across all four views and SofaSpace-Hard.
Among the remaining models, DeepSeek and Kimi achieve identical S accuracy
but differ sharply on G, while Grok attains comparable S performance yet
ranks second on both construction views.
Gemini, despite only 12.00\% on S, is the only non-Astra model with nonzero
R strict success, reaching 9.33\%.
Thus, direct planning accuracy does not characterize performance on
reconstruction or environment construction.

SofaSpace-Hard further separates the evaluated configurations.
Astra achieves 90.64\% strict accuracy over all 534 tasks, followed by Claude at 47.19\%, while the remaining configurations score below 15\%.
The aggregate score, however, masks substantial variation across task types.
Astra reaches 92.81\% on planning and 95.24\% on trajectory-validity judgments but only 26.32\% on the 19 feasibility judgments.
Claude reaches 45.03\%, 92.86\%, and 0.00\% on the same three task types, while Grok reaches 2.11\%, 66.67\%, and 5.26\%, respectively.
The remaining configurations fail all 19 feasibility judgments.

We report mechanism-conditioned performance in Table~\ref{tab:mechanism_composition}.
The SofaSpace test contains 180 single-mechanism and 120 two-mechanism scenes, with no three-way compositions.
Because C01 appears only in C01+C03, C01+C06, and C01+C07, we report exact combinations rather than treating a C01 marginal as an independent category.
Several mechanism-specific differences are pronounced: C04 strongly separates the evaluated configurations, although this group coincides with the paired-cavity body/family holdout and therefore does not isolate concavity reasoning, whereas C06 and C07 are substantially easier for several models.
SofaSpace-Hard covers all seven mechanisms and contains substantially more composed tasks; among weaker configurations, performance generally decreases from single- to two- and three-mechanism compositions, while Astra exhibits the opposite aggregate pattern.
Because these groups differ in family, task type, and mechanism composition, we treat the composition differences as descriptive rather than as a causal effect of adding mechanisms.

\subsection{Where Does Motion Planning Fail}
\label{sec:planning_analysis}

Strict S accuracy does not identify whether a failure occurs in response generation, endpoint specification, submitted configurations, or continuous execution.
We therefore decompose the 180 static SofaSpace test instances into six increasingly strict stages: response completion, parseability, endpoint agreement, submitted-pose validity, continuous-motion validity, and full task validity.

\begin{figure}[t]
    \centering
    \vspace{-.1in}
    \includegraphics[width=\columnwidth]{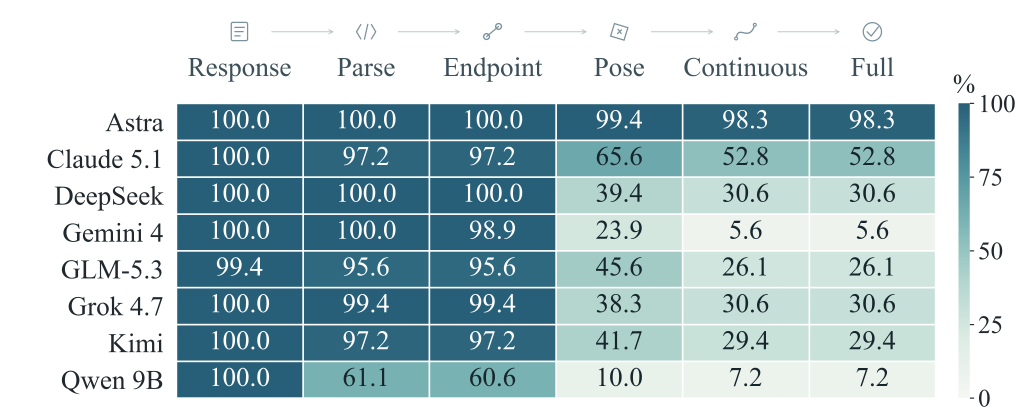}
    \vspace{-.25in}
    \caption{\textbf{Stage-wise validity of SofaSpace motion plans.}
    We progressively require a complete response, parseable trajectory, correct endpoints, valid submitted poses, valid continuous interpolation, and satisfaction of all frozen constraints.
    Values are percentages over the 180 static SofaSpace S instances.}
    \label{fig:planning_diagnostics}
    \vspace{-.1in}
\end{figure}

\begin{figure*}[t]
    \centering
    \includegraphics[width=0.95\textwidth]{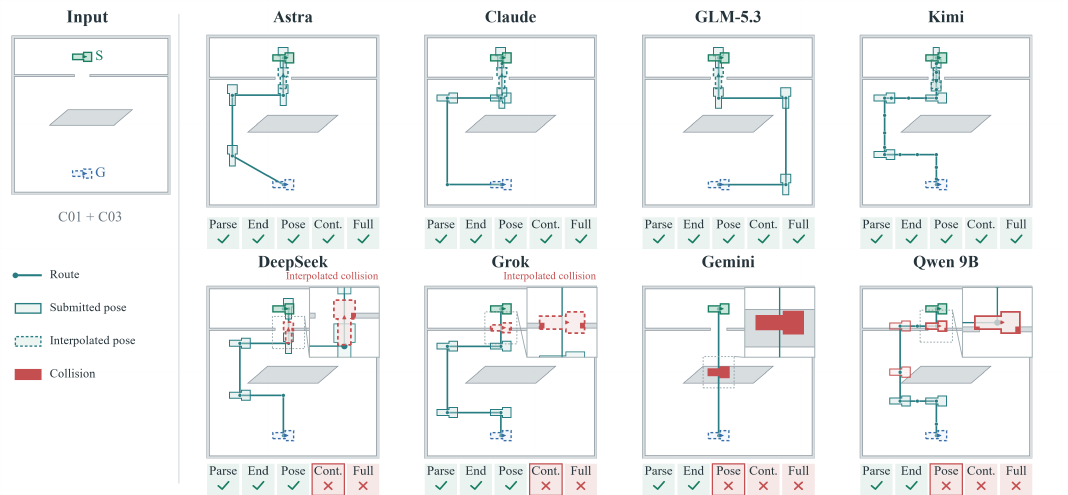}
    \vspace{-.1in}
    \caption{\textbf{Different failure stages on the same SofaSpace S instance.}
    DeepSeek and Grok fail during continuous interpolation despite valid submitted poses, whereas Gemini and Qwen already contain invalid poses.
Indicators denote parsing, endpoint, pose, continuous, and full validity; insets enlarge verified collision regions.}
    \label{fig:qualitative_model_comparison}
    \vspace{-.1in}
\end{figure*}

As shown in Fig.~\ref{fig:planning_diagnostics}, endpoint correctness is not the dominant bottleneck.
DeepSeek, for example, drops from 100.0\% endpoint agreement to 39.4\% submitted-pose validity and 30.6\% continuous validity, while Claude and Kimi exhibit the same layered degradation.
Thus, correct endpoints and individually valid submitted poses do not establish an executable continuous motion.
Representative failures at different stages are shown in Fig.~\ref{fig:qualitative_model_comparison}.

\begin{table}[t]
    \centering
    \caption{\textbf{Oracle-geometry control on 150 paired SofaSpace S scenes.}
    All models are evaluated on the same stratified subset under identical reasoning settings, output contracts, and strict motion verification.
    $\Delta$ denotes the absolute accuracy change in percentage points, and Recovery denotes the fraction of image-only failures that become successful with oracle geometry.}
    \label{tab:oracle_geometry}
    \footnotesize
    \renewcommand{\arraystretch}{1.08}
    \setlength{\tabcolsep}{0pt}
    \begin{tabularx}{\columnwidth}{
        @{}
        >{\raggedright\arraybackslash}p{0.2\columnwidth}
        >{\centering\arraybackslash}X
        >{\centering\arraybackslash}X
        >{\centering\arraybackslash}X
        >{\centering\arraybackslash}X
        @{}
    }
        \toprule
        Model & Image & + Oracle Geo. & $\Delta$ & Recovery \\
        \midrule
        Claude 5.1 & 66.67 & \textbf{100.00} & \textbf{+33.33} & \textbf{100.0} \\
        DeepSeek & 45.33 & \textbf{76.67} & \textbf{+31.33} & 57.3 \\
        Gemini 4 & 12.00 & 13.33 & +1.33 & 11.4 \\
        \bottomrule
    \end{tabularx}
    \vspace{-.1in}
\end{table}

\noindent\textbf{Oracle geometry control.}
To examine how planning depends on recovering precise geometry, we provide exact scene geometry to Claude, DeepSeek, and Gemini on the same 150 SofaSpace S scenes.
The image, task requirements, reasoning setting, output contract, and strict motion verifier remain unchanged, and no reference trajectory or intermediate motion information is provided.
In Table~\ref{tab:oracle_geometry}, we observe large gains for Claude and DeepSeek, whereas Gemini changes marginally, with 15 recoveries offset by 13 regressions; the paired difference is not significant (McNemar $p=0.851$).
These results indicate that precise geometry can remove a substantial bottleneck for some models but does not by itself resolve continuous motion planning.
This control measures the benefit of providing exact geometric representation rather than isolating visual perception as a causal factor.

\begin{table}[t]
    \centering
    \caption{\textbf{Reconstruction and motion-semantic fidelity on the 300 SofaSpace test scenes.}
    Cov. denotes valid-layout coverage; SA and GC denote structural agreement and strict geometry correctness.
    FFMR and VMRR are computed over decided probes.
    $\mathrm{Err}\mid\mathrm{GC}$ counts GC scenes with any decided probe disagreement; ``--'' indicates that no scene satisfies GC.}
    \label{tab:r_fidelity}
    \vspace{-.1in}
    \footnotesize
    \renewcommand{\arraystretch}{1.08}
    \setlength{\tabcolsep}{3pt}
    \begin{tabular}{lrrrrrr}
        \toprule
        Model & Cov. & SA $\uparrow$ & GC $\uparrow$ &
        FFMR $\downarrow$ & VMRR $\downarrow$ &
        $\mathrm{Err}\mid\mathrm{GC}$ $\downarrow$ \\
        \midrule
        Astra
        & 289/300 & 93.67 & 80/300
        & 2.39 & 0.86 & 0/80 \\

        Claude 5.1
        & 19/300 & 6.33 & 0/300
        & 1.69 & 9.26 & -- \\

        DeepSeek
        & 277/300 & 73.67 & 0/300
        & 13.14 & 41.90 & -- \\

        Gemini 4
        & 293/300 & 97.33 & 28/300
        & 0.20 & 6.26 & 0/28 \\

        GLM-5.3
        & 130/300 & 26.67 & 0/300
        & 21.62 & 37.84 & -- \\

        Grok 4.7
        & 288/300 & 70.00 & 0/300
        & 21.40 & 33.18 & -- \\

        Kimi
        & 100/300 & 15.00 & 0/300
        & 36.26 & 34.52 & -- \\

        Qwen 9B
        & 45/300 & 10.00 & 0/300
        & 25.35 & 24.63 & -- \\
        \bottomrule
    \end{tabular}
    \vspace{-.1in}
\end{table}

\subsection{Does Reconstruction Preserve Motion}
\label{sec:reconstruction_analysis}

A reconstruction can recover coarse environment structure while changing the rigid-body motions the environment permits.
We therefore evaluate R at three levels: strict geometry correctness
(GC; reported as R Strict in Table~\ref{tab:main_results}),
coarse structural agreement (SA), and motion-semantic fidelity.
SA requires agreement in free-space component count, hole count, gate count, and gate-interval structure, while GC follows the frozen R acceptance contract detailed in Appendix A.
For motion semantics, we use a fixed, prediction-independent bank of 2,616 probes from the reference environments, including 1,569 valid and 1,047 violated probes.
The false-feasible motion rate (FFMR) measures reference-invalid probes incorrectly admitted by a reconstruction, while the valid-motion rejection rate (VMRR) measures reference-valid probes incorrectly rejected.

\begin{figure}[t]
    \centering
    \includegraphics[width=\columnwidth]{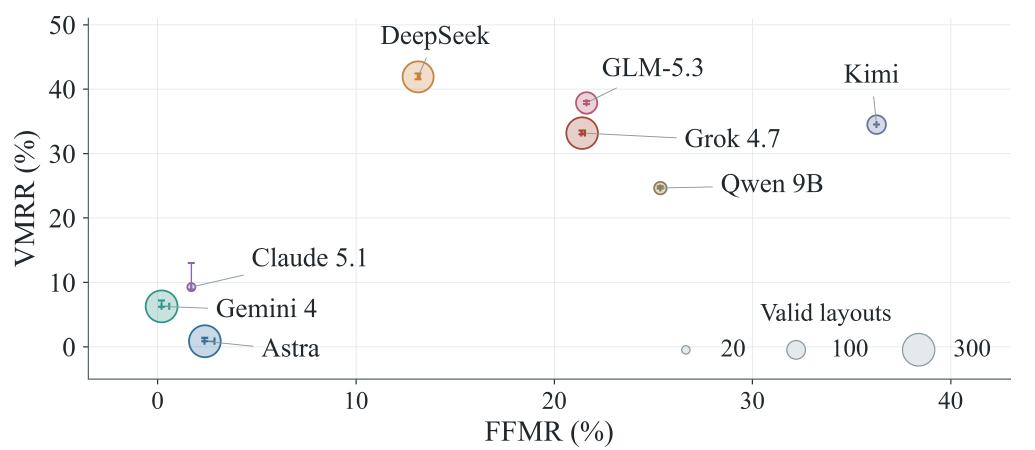}
    \vspace{-.25in}
    \caption{\textbf{Motion-semantic errors in reconstructed environments.}
    FFMR and VMRR measure permissive and restrictive reconstruction errors, respectively.
    Marker area indicates the number of valid reconstructed layouts.}
    \label{fig:reconstruction_diagnostics}
    \vspace{-.1in}
\end{figure}

In Table~\ref{tab:r_fidelity} and Fig.~\ref{fig:reconstruction_diagnostics}, we observe that coarse structural agreement alone does not determine motion-semantic fidelity.
DeepSeek reaches 73.67\% SA but rejects 41.90\% of decided reference-valid probes, while Gemini reaches 97.33\% SA yet rejects 6.26\%.
The stricter GC analysis sharpens this distinction: Astra and Gemini satisfy the full geometry criteria on 80 and 28 scenes, respectively, and none of these geometry-correct reconstructions disagrees on the decided fixed probes.
Among geometry-incorrect scenes with decided probes, motion-semantic disagreements occur in 22/209 Astra scenes and 41/265 Gemini scenes.
The remaining models have no reconstructions satisfying the full GC contract, so the corresponding conditional comparison is unavailable.
These results distinguish coarse structural recovery from precise geometric reconstruction; the fixed probe bank evaluates selected transport relations rather than complete equivalence over configuration space.

\begin{figure}[t]
    \centering
    \includegraphics[width=\columnwidth]{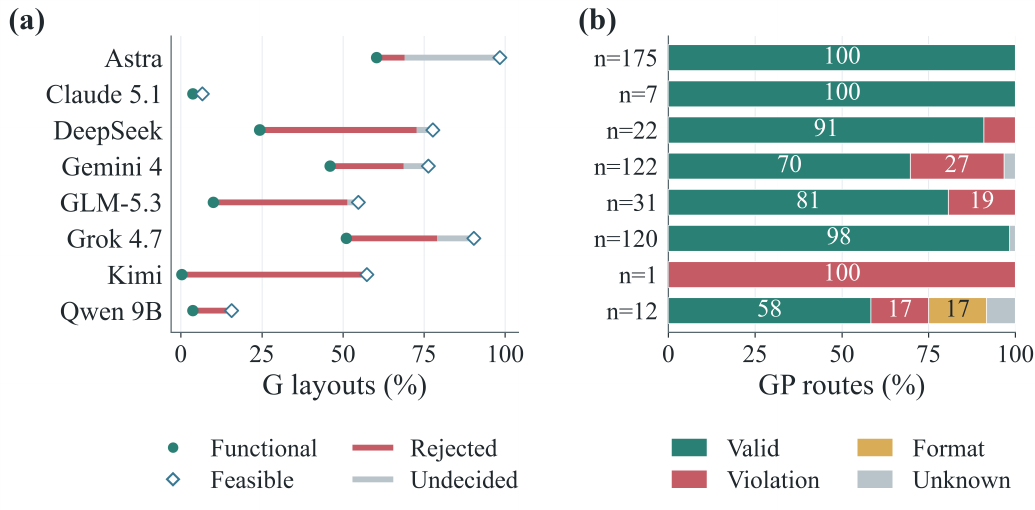}
    \vspace{-.25in}
    \caption{\textbf{Feasibility, functional compliance, and self-execution.}
    (a) G: diamonds denote verified feasible layouts and circles denote layouts satisfying all evaluated functional requirements.
    (b) GP: route outcomes are conditioned on functionally valid generated layouts; $n$ denotes the corresponding conditional sample size.}
    \label{fig:design_diagnostics}
    \vspace{-.1in}
\end{figure}

\subsection{From Feasible Design to Executable Action}
\label{sec:g_gp_analysis}

Environment construction introduces three levels of correctness: geometric feasibility, functional compliance with the prescribed transport requirements, and self-execution in GP.
A feasible layout need not satisfy the required transport behavior, while GP additionally requires a valid motion inside the generated environment.
As shown in Fig.~\ref{fig:design_diagnostics}, geometric feasibility does not imply functional compliance: Astra retains 181 of 295 feasible layouts after functional checking, while Kimi retains only 1 of 172.
Conditional GP evaluation further separates functional design from self-execution.
Astra executes valid routes for all 175 functionally valid layouts, whereas Gemini succeeds on 85 of 122 and Kimi fails on its only functionally valid layout.
Thus, geometric feasibility, functional compliance, and self-execution constitute distinct capability layers, although the size of the conditional execution gap varies across models.

\section{Conclusion}

We introduce SofaSpace, a benchmark for executable visual spatial reasoning through continuous rigid-body transport and environment construction. Across planning, reconstruction, and design, our experiments reveal layered gaps between configuration-level correctness and executable motion, between coarse structural agreement and motion-semantic fidelity, and among geometric feasibility, functional compliance, and self-execution. These results motivate evaluating spatial intelligence through the constraints that make motion possible, while richer geometries, perceptual uncertainty, and embodied execution remain directions for future work.

\section*{Acknowledgments}
The authors used OpenAI ChatGPT and Codex for manuscript drafting and editing, reviewer-style feedback, code, experimental analysis, and figure preparation.
\FloatBarrier
{\small
\bibliographystyle{ieeenat_fullname}
\bibliography{references}
}
\clearpage
\twocolumn[\begin{center}\Large\bfseries Supplementary Material\end{center}\vspace{0.5em}]
\captionsetup[figure]{skip=7pt}
\appendix
\setcounter{figure}{0}
\setcounter{table}{0}
\setcounter{equation}{0}
\renewcommand{\thefigure}{S\arabic{figure}}
\renewcommand{\thetable}{S\arabic{table}}
\renewcommand{\theequation}{S\arabic{equation}}
\ifdefined\theHfigure
  \renewcommand{\theHfigure}{supp.figure.\arabic{figure}}
\fi
\ifdefined\theHtable
  \renewcommand{\theHtable}{supp.table.\arabic{table}}
\fi
\ifdefined\theHequation
  \renewcommand{\theHequation}{supp.equation.\arabic{equation}}
\fi

\section{Detailed Task Specifications and Scoring}
\label{app:task_specs}

\subsection{Motion Rules for the Main S Task}
\label{app:motion_semantics}

Coordinates are expressed in the instance's planar world, with $+x$ pointing
right and $+y$ up.  A pose contains \texttt{x}, \texttt{y}, and
\texttt{theta\_deg}; the angle is counterclockwise about the body's supplied
reference point.  Body vertices use local coordinates.  Between consecutive
poses, position and the explicitly unwrapped angle interpolate linearly.  Thus,
$135^\circ$ to $-135^\circ$ specifies a $-270^\circ$ rotation.  Modulo-$360^\circ$
comparison is used only for endpoint orientation error.  Collision and clearance
checks concern the complete body during each segment, including pairwise
occupancy for multi-body tasks.

Static S answers contain a \texttt{poses} array.  Timed and multi-body answers
contain a \texttt{frames} array; every frame has a timestamp \texttt{t} and a
\texttt{poses} dictionary keyed by body identifier.  Time starts at zero and
strictly increases.  Repeating a pose at a later time expresses waiting.  Linear
and angular speed limits and deadlines are instance-specific.  A timed gate is
open throughout each specified inclusive interval and otherwise contributes
obstacle occupancy; the entire body must clear the gate while it is open.

\subsection{Public Inputs and Output Schemas}
\label{app:public_interfaces}

S receives a complete scene image and task instructions specifying endpoints
and motion rules, and returns a complete motion.  R receives the scene image
plus explicit fixed-body and endpoint context, and returns reconstructed free
space and gates.  G and GP receive fixed bodies, endpoints, and functional requirements.
Any layout satisfying them is admissible. G returns a layout. GP also returns a motion in it.  Reference layouts,
author motions, construction certificates, mechanism labels, and validation
witnesses are not part of the model input.

R, G, and GP encode free space as a valid nonempty GeoJSON Polygon or
MultiPolygon inside the world, with the complement defining fixed obstacles.
Each gate contains polygon geometry and its open intervals.  Supplied bodies,
endpoints, and motion limits cannot be overwritten.  Invalid geometry is
rejected rather than automatically repaired.

\subsection{Acceptance Criteria}
\label{app:acceptance}

Table~\ref{tab:app_contracts} summarizes the numerical criteria used by the
main SofaSpace views.  R strict success is exactly the strict geometry
correctness (GC) metric reported in the main paper: all R requirements in the
table must pass simultaneously.  Structural agreement (SA) is a separate,
coarser diagnostic and is not the primary R success criterion.

\begin{table*}[t]
    \centering
    \caption{\textbf{Motion and reconstruction thresholds.}
    S tolerances are public task requirements.  R strict success requires all
    listed R criteria.  G/GP clearances are instance-specific and are preserved
    during scoring.}
    \label{tab:app_contracts}
    \small
    \setlength{\tabcolsep}{5pt}
    \renewcommand{\arraystretch}{1.0}
    \begin{tabularx}{\textwidth}{@{}l>{\raggedright\arraybackslash}X
        >{\raggedright\arraybackslash}X@{}}
        \toprule
        Scope & Quantity & Requirement \\
        \midrule
        S & Route size; initial state
        & 2--64 poses or frames; start at the supplied initial poses \\
        S & Clearance; final position; final orientation
        & At least $0.01$; position error at most $0.10$; orientation error at
          most $2^\circ$ modulo $360^\circ$ \\
        R & Free-space area and boundary
        & Symmetric-difference area / world area $\leq 0.002$; boundary
          Hausdorff distance $\leq 0.05$ \\
        R & Topology summaries
        & Equal free-space component count and equal total hole count \\
        R & Gates and schedules
        & Equal gate count; matched gate Hausdorff distance $\leq 0.05$;
          equal interval counts and corresponding endpoint errors $\leq 0.02$ \\
        G / GP & Complete-motion clearance
        & The instance-specific clearance required by the design task \\
        GP & Submitted-route endpoints
        & Fixed endpoints with the route checker's numerical endpoint tolerance \\
        \bottomrule
    \end{tabularx}
\end{table*}

S succeeds when its submitted motion satisfies the applicable geometric,
temporal, endpoint, and multi-body constraints.  For R, gate matching does not
require identical list order; component and hole counts are summaries rather
than complete topological equivalence.  G requires legal geometry, a verified
complete-motion witness at the required clearance, and every evaluated public
functional condition.  Alternative layouts are admissible.  In GP, a valid
submitted motion can provide the feasibility witness, but joint success also
requires all layout conditions.  Rejecting a submitted GP route does not by
itself establish that the generated layout is unsolvable.  An unresolved
feasibility search remains \texttt{unknown}; it is neither a verified success
nor evidence of impossibility.

For tasks requiring two detours around a single island, the verifier checks for
a complete rigid-body motion between the same start and goal poses on each
side of the island, at the required clearance. For tasks with lower-branch
orientation constraints, it verifies that every complete lower-branch motion
must cross each specified port and that the body cannot occupy the port at any
orientation within $3^\circ$ of either $0^\circ$ or $180^\circ$. The verifier certifies mandatory passage by checking that the start and goal become disconnected when crossings through the port are disallowed, and excludes the forbidden headings using conservative displacement bounds over the port's position and orientation intervals.

\section{Dataset Construction, Mechanisms, and Verification}
\label{app:construction}

\subsection{Scene Families and Mechanism Coverage}

SofaSpace combines mechanism-driven parent families with parameterized
variants.  A parent family denotes a shared structural or generative template;
changes in dimensions, reflection, endpoint pose, or mechanism labels do not by
themselves establish a new family.  The 3,000 physical scenes are split by parent family. Registered parent families are disjoint
across train, development, and test splits, with derivatives grouped together.

Mechanism tags identify the construction requirements instantiated by each scene.
Table~\ref{tab:app_mechanism_coverage} gives their counts across splits.
A scene with multiple tags contributes to each corresponding row.

\begin{table}[t]
    \centering
    \caption{\textbf{Recorded mechanism coverage in SofaSpace.}
    Counts are physical scenes carrying each tag and may overlap.  A further
    300 training scenes are generic/untagged.}
    \label{tab:app_mechanism_coverage}
    \small
    \setlength{\tabcolsep}{5pt}
    \begin{tabular}{lrrr}
        \toprule
        Tag & Train & Dev & Test \\
        \midrule
        C01 & 665 & 50 & 120 \\
        C02 & 545 & 0  & 0 \\
        C03 & 320 & 25 & 120 \\
        C04 & 320 & 25 & 60 \\
        C05 & 585 & 25 & 0 \\
        C06 & 275 & 25 & 60 \\
        C07 & 290 & 25 & 60 \\
        Generic & 300 & 0 & 0 \\
        \bottomrule
    \end{tabular}
\end{table}

The mutually exclusive exact compositions of the 300-scene test split are
C03 (60), C04 (60), C06 (30), C07 (30), C01+C03 (60), C01+C06 (30), and
C01+C07 (30).  C01 occurs in the three paired compositions listed above. The 60 C04 scenes form the paired-cavity body-and-family holdout.

\begin{figure*}[!tp]
\centering
\includegraphics[width=\textwidth]{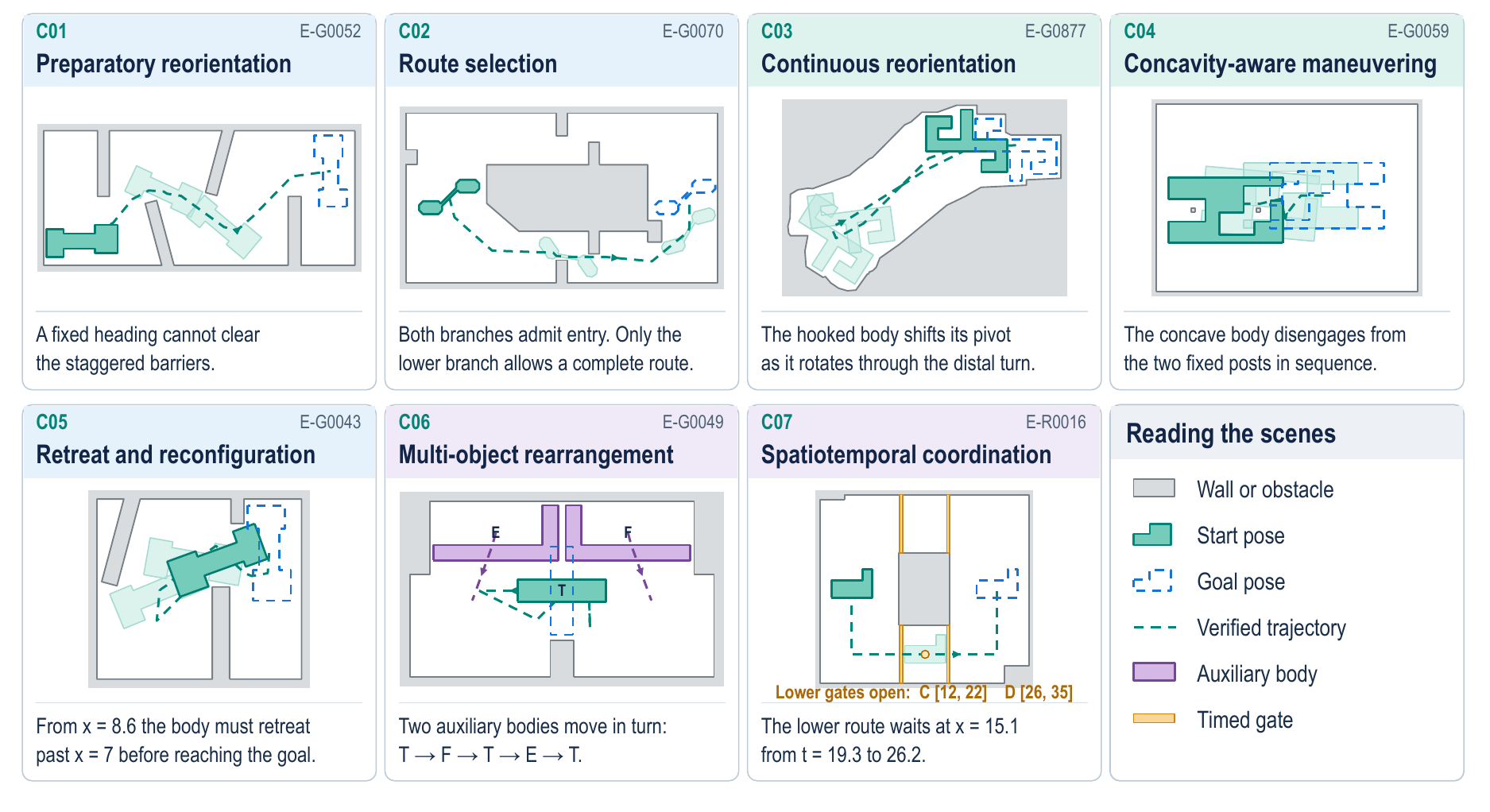}
\caption{\textbf{Real single-mechanism examples.} Each selected Hard task carries one recorded mechanism. The overlaid author witnesses were verified separately and are omitted from public task inputs.}
\label{fig:app_hard_examples}
\end{figure*}

Figure~\ref{fig:app_hard_examples} illustrates the seven mechanisms using single-tagged tasks from SofaSpace-Hard.

\subsection{Continuous Motion Verification and Constraint Evidence}
\label{app:verification_details}

Feasibility is supported by complete motion witnesses under the interpolation
contract.  For a linearly interpolated pose segment, let $\rho$ be the maximum
distance from the body's reference point to its vertices.  A material-point
displacement bound is
\begin{equation}
    L=\lVert\Delta\mathbf{x}\rVert_2+\rho\lvert\Delta\theta\rvert,
    \qquad \Delta\theta\ \text{in radians}.
    \label{eq:app_motion_bound}
\end{equation}
With $n$ equal intervals, displacement to the nearest sample is at most
$L/(2n)$ by Eq.~\ref{eq:app_motion_bound}.  The verifier combines sampled clearance with this bound,
checks containment, and refines inconclusive segments.  Multi-body and timed
trajectories additionally check body--body separation, gate occupancy, speeds,
waiting, deadlines, and endpoints.  The verifier uses floating-point geometry with conservative displacement bounds and returns unknown when refinement is inconclusive.

For SS3-TR1021, the author motion clears the original bend (clearance bound
$0.8326$, rounded). Removing the bend walls admits a $315^\circ$
fixed-heading trajectory with the same body and endpoints (clearance bound
$1.2100$, rounded). Restoring the walls puts $5.1648$ square units of the body
outside free space at a recorded pose on that trajectory.
This replay rules out one specific shortcut. A separate certificate rules out
an entire strategy class for this instance: even an outer approximation of
free space with headings restricted to $315^\circ\!\pm\!3^\circ$ disconnects
start and goal. Every complete motion must therefore leave that band.
The seven witnesses in Fig.~\ref{fig:app_hard_examples} establish feasibility
of their respective instances.

\begin{figure*}[!tp]
\centering
\includegraphics[width=\textwidth]{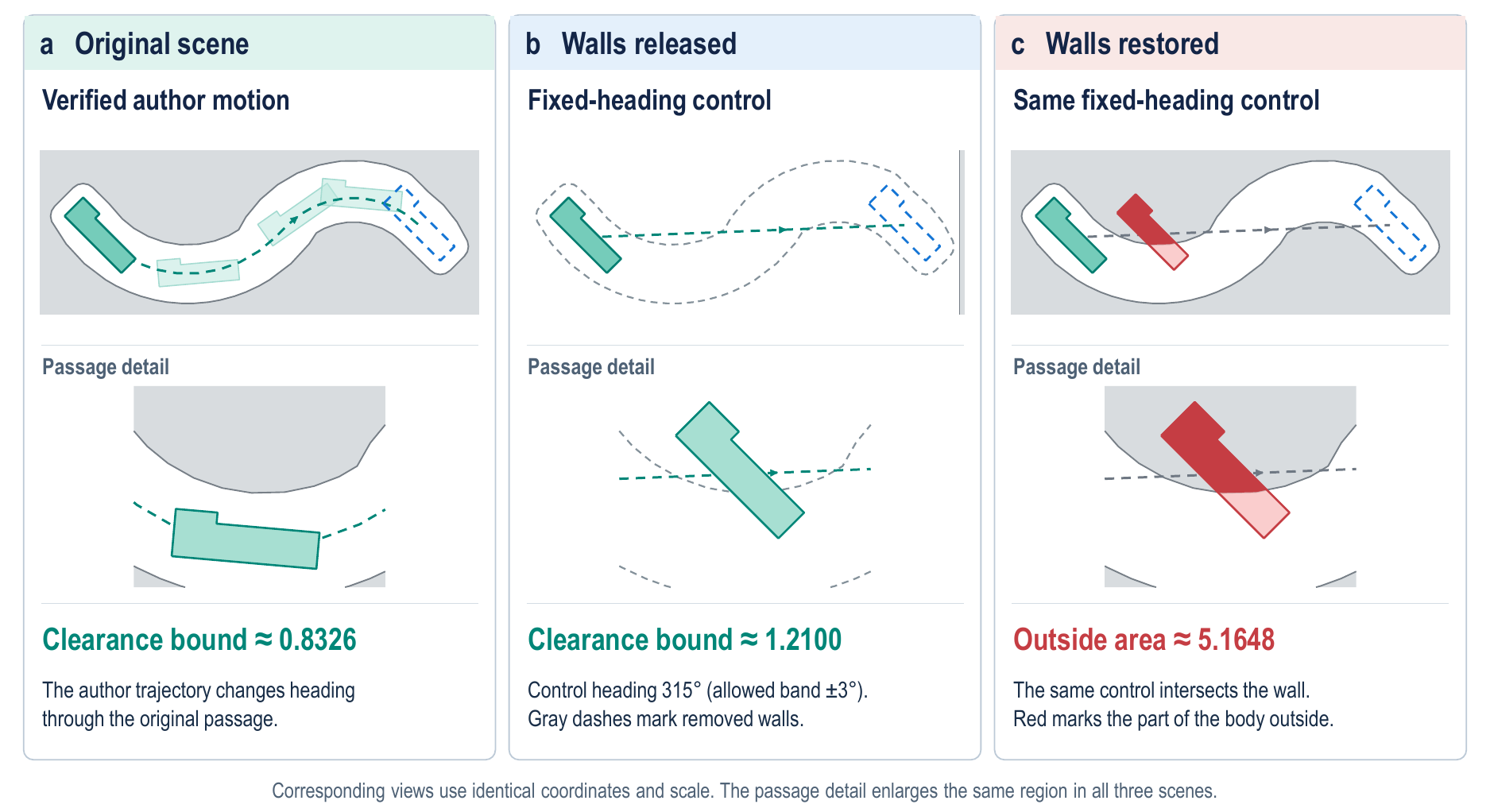}
\caption{\textbf{Constraint-release control for SS3-TR1021.} The author and control motions use the same rigid body and endpoints. A fixed-heading trajectory clears the relaxed passage. Restoring the original walls makes that same trajectory collide.}
\label{fig:app_constraint_release}
\end{figure*}

\onecolumn
\noindent\textbf{Two checks on Hard examples in Fig.~\ref{fig:app_hard_examples}.}
In E-G0049 (C06), each parked auxiliary body overlaps a region occupied
by the target at every admissible goal pose. Thus both bodies must move,
although the certificate leaves their order open. In E-R0016 (C07), a
trajectory through the upper passage clears the first gate and completes
when the downstream window is relaxed. Replaying it under the original
schedule produces a closed-gate collision at gate B on segment 6.
The lower passage admits a complete timed motion. The failed replay rules
out the tested upper-passage schedule, while allowing retreat to the lower passage.

\subsection{Quality Control and Split Integrity}

Admission to the 3,000-scene SofaSpace dataset combines geometric legality, complete-motion feasibility,
mechanism-specific evidence, and visual review.  View eligibility is
interface-specific because a scene can have a resolved certificate for one
view but not another.  The final dataset therefore contains 2,600/100/300
train/dev/test scenes, with G eligibility 2,564/100/300 and GP eligibility
2,599/100/300.

Each of the 300 test scenes supplies four aligned task inputs (1,200 instances).
Cross-view comparisons therefore share the same 300 physical scenes.
The test split contains 180 unseen-family scenes with familiar body
constructions, 60 scenes with a held-out gate-dependency structure, and 60
scenes jointly holding out the paired-cavity body construction and its
associated family. Cross-split exact-duplicate checks cover records, layouts, fixed contexts, and rendered images.

\FloatBarrier

\section{Experimental Protocol and Reproducibility}
\label{app:experimental_protocol}

\subsection{Inputs and Model Configurations}

In the main evaluation, each model receives only the public image, task prompt, and fixed context for
the requested view.  No demonstrations, private reference layouts, author
trajectories, or validation certificates are supplied.  Table~\ref{tab:app_model_configs}
lists the eight evaluated configurations and requested reasoning modes.

\par\medskip\noindent\begin{minipage}{\textwidth}
    \centering
    \captionof{table}{\textbf{Evaluated model configurations.}
    API identifiers are the recorded request strings.  The local Qwen row
    denotes the pinned checkpoint used in the experiments.}
    \label{tab:app_model_configs}
    \small
    \setlength{\tabcolsep}{5pt}
    \begin{tabularx}{\textwidth}{@{}l>{\raggedright\arraybackslash}X
        >{\raggedright\arraybackslash}X@{}}
        \toprule
        Model & Requested model / checkpoint & Requested reasoning setting \\
        \midrule
        GPT-6 Astra & \texttt{gpt-6-astra} & xhigh \\
        Claude Fable 5.1 & \texttt{claude-fable-5-1} & adaptive thinking; maximum effort \\
        DeepSeek V4.1 Flash & \texttt{deepseek-v4.1-flash} & thinking enabled; maximum effort \\
        Gemini 4 Argon & \texttt{gemini-4-argon-high} & high-reasoning endpoint \\
        GLM-5.3-Flash & \texttt{glm-5.3-flash} & thinking enabled; maximum effort \\
        Grok 4.7 & \texttt{grok-4.7} & xhigh \\
        Kimi K3 & \texttt{kimi-k3} & maximum effort \\
        Qwen3.5-9B & \texttt{Qwen/Qwen3.5-9B} & thinking enabled \\
        \bottomrule
    \end{tabularx}
\end{minipage}\par\medskip

For Qwen3.5-9B we use the pinned original checkpoint with bfloat16, batch size
one, thinking enabled, and vLLM 0.19.1.  Sampling uses temperature 1,
top-$p=0.95$, top-$k=20$, min-$p=0$, presence penalty 1.5, and repetition
penalty 1.0.

\subsection{Response Selection and Scoring}

Requests interrupted before an answer was returned could be retried without
counting as a model response. An answer truncated at the output limit counted
as a failed first response and received zero credit unless a retry supplied the
final answer used for scoring. Relative to the first-response scores, retries
added seven passes for DeepSeek on S, one for Kimi on S, and two for Claude on
Hard planning.

\subsection{Feasibility-Search Reliability for G}
\label{app:g_search_reliability}

We test the feasibility search on known-feasible layouts and withhold their verified trajectories from the search. The reference cohort contains all 300 test layouts.
The audit reported in the main paper uses 942 GP layouts with independently
verified submitted routes from seven configurations (Astra, Claude, DeepSeek,
Gemini, GLM-5.3, Kimi, and Qwen).  Table~\ref{tab:app_g_reliability}
shows the recovery results.

\par\medskip\noindent\begin{minipage}{\textwidth}
    \centering
    \captionof{table}{\textbf{Feasibility-search reliability.} (a) Recovery on known-feasible layouts.
    The last two columns count layouts for which the search found a complete motion under the default (1$\times$) and doubled (2$\times$) budgets.}
    \label{tab:app_g_reliability}
    \small
    \setlength{\tabcolsep}{3.5pt}
    \begin{tabular}{lrrr}
        \toprule
        Known-feasible source & $N$ & 1$\times$ solved & 2$\times$ solved \\
        \midrule
        Reference layouts & 300 & 300 & 300 \\
        GP layouts in the main-paper audit & 942 & 925 & 925 \\
        Unique GP search inputs & 705 & 688 & 688 \\
        \bottomrule
    \end{tabular}
\end{minipage}\par\medskip

The default audit allocates 180 seconds to feasibility search; the doubled
condition allocates 360 seconds and also doubles lattice-node caps, repair
attempts, staging-chain caps, and nested search-time ceilings.  The active
search is deterministic with seed 0.  The two budgets yield identical outcomes for every layout. Search recovers
925 of 942 known-feasible GP layouts (98.20\%). The remaining 17 searches
return unknown under both budgets.

\par\medskip\noindent\begin{minipage}{\textwidth}
\centering
\captionof*{table}{\textbf{Table S4 (continued).} (b) Complete-motion search outcomes on actual G submissions (300 per model).
The last four columns partition the search-unresolved layouts by their other functional checks.}
\small\setlength{\tabcolsep}{6pt}
\begin{tabular}{lrrrrr}
\toprule
Model & Search unresolved & Violation & Undecided & Check error & All others passed \\
\midrule
Astra & 5 & 3 & 1 & 1 & 0 \\
Claude & 0 & 0 & 0 & 0 & 0 \\
DeepSeek & 12 & 8 & 1 & 0 & 3 \\
Gemini & 22 & 13 & 4 & 0 & 5 \\
GLM-5.3 & 20 & 9 & 6 & 0 & 5 \\
Grok & 17 & 16 & 0 & 0 & 1 \\
Kimi & 3 & 3 & 0 & 0 & 0 \\
Qwen 9B & 3 & 3 & 0 & 0 & 0 \\
\midrule
Total & 82 & 55 & 12 & 1 & 14 \\
\bottomrule
\end{tabular}
\end{minipage}\par\smallskip
Among 2,400 G submissions from eight models, 82 returned an unresolved
complete-motion search. Table S4(b) partitions these submissions by the other
functional checks. Of the 82 layouts, 14 passed every other evaluated condition.
Holding those checks fixed, even if a feasible motion were found for all 14,
strict G accuracy would increase by at most 5/300 (1.67 percentage points)
for any one model.

\section{Additional Diagnostic Results}
\label{app:diagnostics}

\subsection{Where Motion Plans Fail}

The main paper reports the six-stage funnel on the 180 static S test scenes.
For each model, we report the fraction of pose-valid submissions that also pass
continuous verification (Table~\ref{tab:app_s_funnel}).

\par\smallskip\noindent\begin{minipage}{\textwidth}
    \centering
    \captionof{table}{\textbf{Stage-wise validity on the 180 static S test scenes.}
    Entries are accepted counts out of 180.  The final column is the fraction
    of pose-valid responses that also pass continuous interpolation.}
    \label{tab:app_s_funnel}
    \renewcommand{\arraystretch}{0.85}
    \small
    \setlength{\tabcolsep}{4pt}
    \begin{tabular}{lrrrrrrr}
        \toprule
        Model & Response & Parse & Endpoint & Pose & Continuous & Full & Cont.$\mid$Pose (\%) \\
        \midrule
        Astra & 180 & 180 & 180 & 179 & 177 & 177 & 98.9 \\
        Claude & 180 & 175 & 175 & 118 & 95 & 95 & 80.5 \\
        DeepSeek & 180 & 180 & 180 & 71 & 55 & 55 & 77.5 \\
        Gemini & 180 & 180 & 178 & 43 & 10 & 10 & 23.3 \\
        GLM-5.3 & 179 & 172 & 172 & 82 & 47 & 47 & 57.3 \\
        Grok & 180 & 179 & 179 & 69 & 55 & 55 & 79.7 \\
        Kimi & 180 & 175 & 175 & 75 & 53 & 53 & 70.7 \\
        Qwen 9B & 180 & 110 & 109 & 18 & 13 & 13 & 72.2 \\
        \bottomrule
    \end{tabular}
\end{minipage}\par\smallskip

\enlargethispage{0.7\baselineskip}
Across the 180 static S test scenes, 655 submissions pass every individual pose check, but only 505 pass continuous verification. Of the 150 submissions that drop out, 145 have a verified collision at an interpolated pose between two consecutive submitted poses. For the remaining five, the frozen verifier returns \texttt{unknown} for continuous clearance. All five nevertheless have a recorded inter-frame sample with clearance below the required 0.01, with zero recorded area outside the free space. These results reveal a specific weakness in rigid-body planning: models can select poses that are individually valid yet fail to connect them into a motion that maintains clearance throughout.

\par\smallskip\noindent\begin{minipage}{\textwidth}
\centering
\addtocounter{table}{-1}
\renewcommand{\thetable}{S\arabic{table}b}
\renewcommand{\theHtable}{supp.figuretable.dynamic}
\captionof{table}{\textbf{Motion outcomes on the 120 dynamic S test scenes.}
Each model plans in 60 multi-body scenes and 60 gated scenes. Each submission has
one final outcome. A collision can appear in several evidence columns.}
\label{tab:app_s_dynamic}
\small
\setlength{\tabcolsep}{3pt}
\begin{tabular*}{\textwidth}{@{\extracolsep{\fill}}lrrrrrrr@{\hspace{9pt}}rrrrr@{}}
\toprule
& \multicolumn{7}{c}{Final outcome (120 per model)}
& \multicolumn{5}{c}{Collision evidence (submissions)} \\
\cmidrule(lr){2-8}\cmidrule(lr){9-13}
Model & Pass & Collision & Speed & Endpoint & Format & Trunc. & Unknown
& \shortstack{Wall at\\frame} & \shortstack{Wall between\\frames}
& \shortstack{Body\\pair} & \shortstack{Closed\\gate} & \shortstack{All frames\\valid$^{*}$} \\
\midrule
Astra & 120 & 0 & 0 & 0 & 0 & 0 & 0 & 0 & 0 & 0 & 0 & 0 \\
Claude & 109 & 4 & 4 & 0 & 0 & 0 & 3 & 1 & 3 & 1 & 2 & 1 \\
DeepSeek & 88 & 31 & 1 & 0 & 0 & 0 & 0 & 19 & 31 & 1 & 1 & 4 \\
Gemini & 26 & 79 & 9 & 6 & 0 & 0 & 0 & 70 & 77 & 24 & 1 & 0 \\
GLM-5.3 & 48 & 30 & 11 & 0 & 30 & 0 & 1 & 19 & 28 & 4 & 8 & 2 \\
Grok & 93 & 27 & 0 & 0 & 0 & 0 & 0 & 13 & 27 & 0 & 0 & 4 \\
Kimi & 90 & 30 & 0 & 0 & 0 & 0 & 0 & 12 & 29 & 0 & 3 & 6 \\
Qwen 9B & 0 & 17 & 50 & 2 & 48 & 3 & 0 & 12 & 14 & 2 & 7 & 0 \\
\midrule
Total & 574 & 218 & 75 & 8 & 78 & 3 & 4 & 146 & 209 & 32 & 22 & 17 \\
\bottomrule
\end{tabular*}
\par\smallskip
\raggedright\footnotesize
$^{*}$Subset of between-frame wall collisions for which all submitted frames satisfy the geometric checks.
\end{minipage}\par\smallskip

Across the 960 dynamic submissions, 218 have a verified collision (Table~\ref{tab:app_s_dynamic}). Of these, 209 collide
with the fixed environment between submitted frames. For at least 17 of the 209,
every submitted frame is geometrically valid, so checking only those frames would
miss the collision. Body-pair and closed-gate counts record overlap between
two bodies and with a gate during a closed interval, respectively.

\par\smallskip\noindent\begin{minipage}{\textwidth}
\textbf{Supervised adaptation on motion planning.}
We fine-tuned Qwen3.5-9B for one epoch on the 2,600 S training scenes. We froze the vision encoder and trained rank-8 LoRA adapters on the language model ($\alpha=16$, dropout 0.05, learning rate $2\times10^{-5}$, effective batch size 8). We supervised the final trajectory JSON and the end-of-sequence token and evaluated the predetermined final checkpoint at step 325.
Under final strict scoring on the 300 S test tasks, Qwen3.5-9B passed 13 (4.33\%, Table~3 of the main paper) and the LoRA-adapted model passed 23 (7.67\%).
\par\medskip\centering
\includegraphics[width=.96\textwidth]{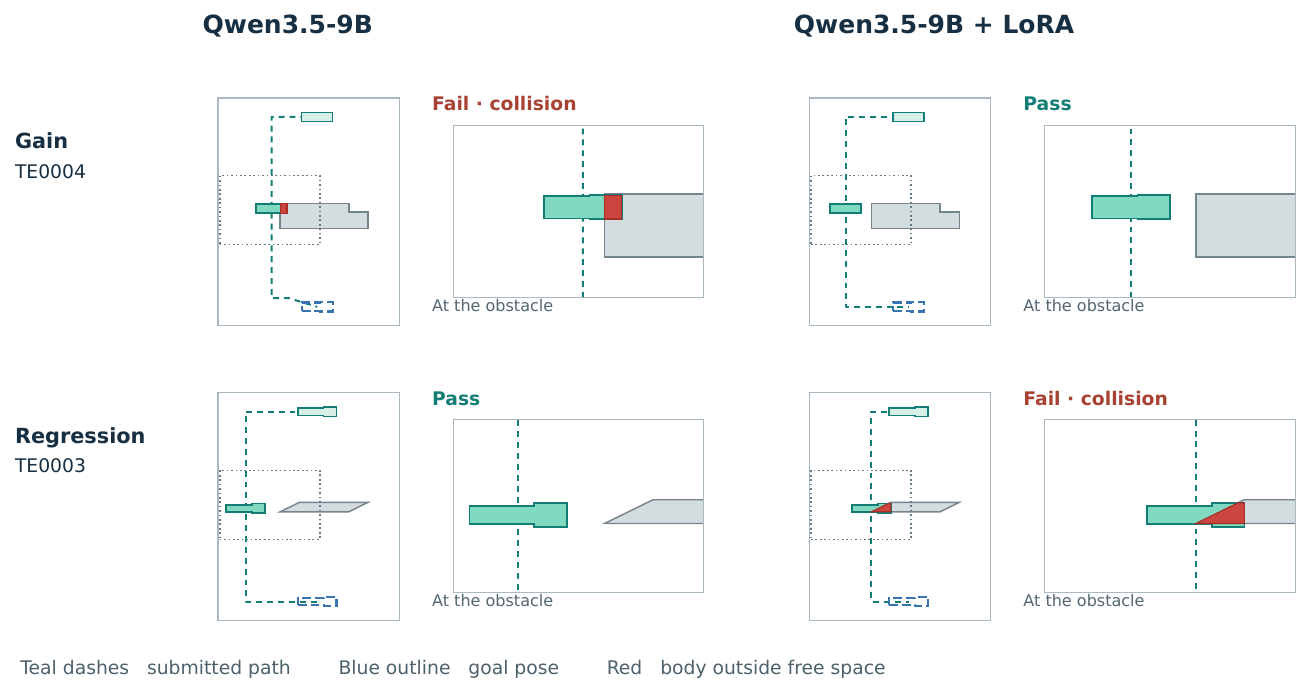}\par\vspace{7pt}
\raggedright\small \textbf{Figure D1. Paired planning trajectories before and after adaptation.}
The close-ups in each row show the body at the same height beside the obstacle. Both examples change under strict scoring.
Among 249 test tasks with complete first responses from both configurations, 13 changed from fail to pass and three from pass to fail after JSON extraction. Twelve of the 13 gains were single-obstacle detours.
\end{minipage}\par\medskip

\subsection{Oracle-Geometry Control}
\label{app:oracle_geometry}

The oracle-geometry control uses the same 150 scenes for Claude, DeepSeek, and
Gemini.  The subset is stratified by exact test composition: C03 (30), C04
(30), C06 (15), C07 (15), C01+C03 (30), C01+C06 (15), and C01+C07 (15).
We add only the exact scene geometry to the original input and keep each model's reasoning setting, output contract, and strict S verifier fixed.

\par\medskip\noindent\begin{minipage}{\columnwidth}
    \centering
    \captionof{table}{\textbf{Paired transitions in the oracle-geometry control.}
    F$\rightarrow$S denotes an image-only failure recovered with oracle
    geometry; S$\rightarrow$F denotes a regression.}
    \label{tab:app_oracle_transitions}
    \small
    \setlength{\tabcolsep}{4pt}
    \begin{tabular}{lrrrr}
        \toprule
        Model & F$\rightarrow$F & F$\rightarrow$S & S$\rightarrow$F & S$\rightarrow$S \\
        \midrule
        Claude & 0 & 50 & 0 & 100 \\
        DeepSeek & 35 & 47 & 0 & 68 \\
        Gemini & 117 & 15 & 13 & 5 \\
        \bottomrule
    \end{tabular}
\end{minipage}\par\medskip

Claude rises from 100/150 to 150/150 and DeepSeek from 68/150 to 115/150,
with no regressions (Table~\ref{tab:app_oracle_transitions}).  Gemini changes from 18/150 to 20/150, with 15 failure-to-success
transitions and 13 success-to-failure transitions (exact McNemar $p=0.851$).
On the 90 static scenes, we identified 37 image-only responses with verified collisions between legal submitted poses. With exact geometry, the models returned valid trajectories in 22 of those cases. In nine of Gemini's 17 such cases, every submitted pose was again legal, but the interpolated motion still collided.

\subsection{Reconstruction Coverage and Motion-Semantic Probes}
\label{app:r_probes}

The fixed R probe bank contains 2,616 prediction-independent motions derived
from reference environments.  Probes are replayed only when the model output is
a valid reconstructed layout.  A decided reference-invalid probe that becomes
valid contributes to FFMR; a decided reference-valid probe that becomes invalid
contributes to VMRR.  Table~\ref{tab:app_r_probe_coverage} reports both rates alongside the number of layouts
eligible for replay, their decided-probe denominators, and the number of unknown
outcomes excluded from those denominators.

Table~\ref{tab:app_sa_conditioned} restricts replay to reconstructions that match the
reference in component count, hole count, and gate structure. For Gemini, 38 of
292 such scenes reject at least one reference-valid motion (85 of 1,492 decided valid probes).

Figure~\ref{fig:app_semantic_examples} traces three fixed probes through the reference and reconstructed layouts.

\newpage
\par\smallskip\noindent
\begin{minipage}[t]{.48\textwidth}
\centering
    \captionof{table}{\textbf{R probe coverage and errors.}
    SA uses all 300 test scenes. Only valid reconstructed layouts enter replay.
    FFMR uses decided reference-invalid probes as its denominator.
    VMRR uses decided reference-valid probes. Unknown replay outcomes are
    excluded from both rates.}
    \label{tab:app_r_probe_coverage}
    \scriptsize
    \setlength{\tabcolsep}{2pt}
    \begin{tabular}{lrrrrrrr}
        \toprule
        Model & \shortstack{Replay\\scenes} & \shortstack{SA\\(\% of 300)} & GC & \shortstack{Dec.\\valid} & \shortstack{Dec.\\invalid} & Unknown & \shortstack{FFMR /\\VMRR (\%)} \\
        \midrule
        Astra & 289 & 93.67 & 80 & 1,504 & 1,006 & 13 & 2.39 / 0.86 \\
        Claude & 19 & 6.33 & 0 & 162 & 59 & 7 & 1.69 / 9.26 \\
        DeepSeek & 277 & 73.67 & 0 & 1,444 & 959 & 14 & 13.14 / 41.90 \\
        Gemini & 293 & 97.33 & 28 & 1,501 & 1,016 & 19 & 0.20 / 6.26 \\
        GLM-5.3 & 130 & 26.67 & 0 & 666 & 444 & 3 & 21.62 / 37.84 \\
        Grok & 288 & 70.00 & 0 & 1,504 & 1,000 & 11 & 21.40 / 33.18 \\
        Kimi & 100 & 15.00 & 0 & 562 & 342 & 0 & 36.26 / 34.52 \\
        Qwen 9B & 45 & 10.00 & 0 & 272 & 142 & 1 & 25.35 / 24.63 \\
        \bottomrule
    \end{tabular}
\end{minipage}\hfill\begin{minipage}[t]{.48\textwidth}
\centering
\addtocounter{table}{-1}
\renewcommand{\thetable}{S\arabic{table}b}
\renewcommand{\theHtable}{supp.table.sa}
\captionof{table}{\textbf{Motion errors in reconstructions with structural agreement (SA).}
Scene counts give the number of SA scenes with at least one error of the indicated type.
Probe counts give errors over decided reference-invalid probes for false admissions and
decided reference-valid probes for false rejections.}
\label{tab:app_sa_conditioned}
\scriptsize
\setlength{\tabcolsep}{2.6pt}
\begin{tabular}{lrrrrr}
\toprule
& & \multicolumn{2}{c}{False admission} & \multicolumn{2}{c}{False rejection} \\
\cmidrule(lr){3-4}\cmidrule(lr){5-6}
Model & SA scenes & Scenes & Probes & Scenes & Probes \\
\midrule
Astra & 281 & 1 & 1/983 & 13 & 13/1,441 \\
Claude & 19 & 1 & 1/59 & 6 & 15/162 \\
DeepSeek & 221 & 10 & 10/741 & 103 & 485/1,081 \\
Gemini & 292 & 2 & 2/1,013 & 38 & 85/1,492 \\
GLM-5.3 & 80 & 6 & 8/260 & 42 & 137/365 \\
Grok & 210 & 11 & 12/694 & 106 & 473/1,026 \\
Kimi & 45 & 8 & 16/138 & 24 & 57/184 \\
Qwen 9B & 30 & 2 & 3/89 & 10 & 30/162 \\
\bottomrule
\end{tabular}
\end{minipage}\par\medskip
\subsection{Conditioning Motion Semantics on Strict Geometry}
\noindent\begin{minipage}[t]{.48\textwidth}

{\small We group reconstructions by whether they satisfy the fixed geometry criteria in Sec.~\ref{app:acceptance} (Table~\ref{tab:app_contracts}) and measure motion-probe disagreement in each group.

No decided motion-semantic disagreement is observed in the geometry-correct
subsets of Astra (0/80 scenes) or Gemini (0/28 scenes).  The corresponding
probe-level conditional rates are also zero: Astra has 0/270 FFMR and 0/303
VMRR events, while Gemini has 0/92 and 0/53.  Table~\ref{tab:app_r_conditional} reports N/A for configurations with no geometry-correct reconstructions.\par}\end{minipage}\hfill\begin{minipage}[t]{.48\textwidth}
\centering
    \captionof{table}{\textbf{Scene-level motion-semantic disagreement conditioned on
    strict geometry correctness.}
    The GC=0 denominator includes scenes with at least one decided replay.
    N/A means that no reconstruction satisfies GC for that model.}
    \label{tab:app_r_conditional}
    \scriptsize
    \setlength{\tabcolsep}{2pt}
    \begin{tabular}{lrrrr}
        \toprule
        Model & GC scenes & \shortstack{Error among\\GC} & \shortstack{Evaluable\\GC=0 scenes} & \shortstack{Error among\\GC=0} \\
        \midrule
        Astra & 80 & 0/80 (0.0\%) & 209 & 22/209 (10.5\%) \\
        Claude & 0 & N/A & 19 & 7/19 (36.8\%) \\
        DeepSeek & 0 & N/A & 277 & 166/277 (59.9\%) \\
        Gemini & 28 & 0/28 (0.0\%) & 265 & 41/265 (15.5\%) \\
        GLM-5.3 & 0 & N/A & 130 & 95/130 (73.1\%) \\
        Grok & 0 & N/A & 288 & 187/288 (64.9\%) \\
        Kimi & 0 & N/A & 100 & 87/100 (87.0\%) \\
        Qwen 9B & 0 & N/A & 45 & 26/45 (57.8\%) \\
        \bottomrule
    \end{tabular}
\end{minipage}
\par\vspace{2pt}
\noindent\begin{minipage}{\textwidth}
\centering
\includegraphics[width=.92\textwidth]{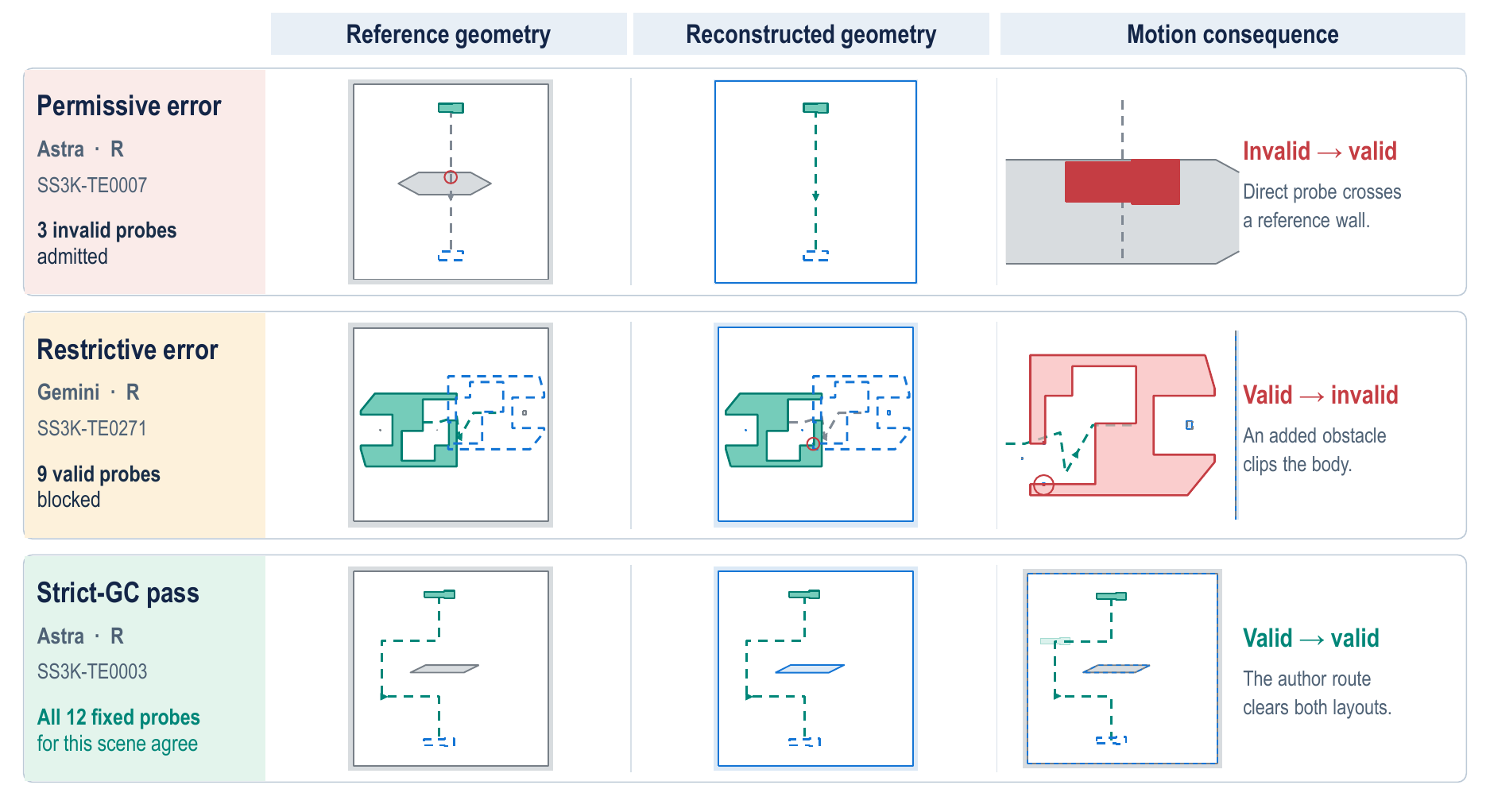}
\captionof{figure}{\textbf{Motion semantics under reconstruction.} A removed obstacle can admit an invalid probe, and an added obstacle can block a valid one. Agreement in the third row covers the twelve fixed probes for that scene.}
\label{fig:app_semantic_examples}
\end{minipage}
\newpage
\subsection{Environment Construction and Self-Execution}

Table~\ref{tab:app_g_gp_layers} gives the counts underlying the construction
analysis.  G feasibility is established by a verified complete motion.  The
functional-valid count is the subset satisfying all evaluated task-specific
functional conditions.  The GP route column is conditioned on a functionally
valid generated layout and therefore has a different denominator from the
300-item joint score.

Across 2,400 G submissions, 1,431 layouts have a verified complete motion;
598 of these also pass all functional checks. Thus 833 witness-bearing
layouts fall short of strict design success. GP uses a separate set of
submissions: among 490 layouts that pass the functional checks, 437
submitted routes pass execution and 53 do not. For Gemini alone, 85 of
122 such routes pass.

Figure~\ref{fig:app_construction_examples} illustrates how the three checks apply to individual generated layouts.

\par\medskip
\noindent\begin{minipage}{\textwidth}
\centering
    \captionof{table}{\textbf{Construction layers for G and GP.}
    GP route validity is evaluated among functionally valid generated layouts.}
    \label{tab:app_g_gp_layers}
    \small
    \setlength{\tabcolsep}{5pt}
    \begin{tabular}{lrrrr}
        \toprule
        Model & G verified feasible & G functional valid & GP functional layouts & GP valid route \\
        \midrule
        Astra & 295 & 181 & 175 & 175/175 \\
        Claude & 20 & 11 & 7 & 7/7 \\
        DeepSeek & 233 & 73 & 22 & 20/22 \\
        Gemini & 229 & 138 & 122 & 85/122 \\
        GLM-5.3 & 164 & 30 & 31 & 25/31 \\
        Grok & 271 & 153 & 120 & 118/120 \\
        Kimi & 172 & 1 & 1 & 0/1 \\
        Qwen 9B & 47 & 11 & 12 & 7/12 \\
        \bottomrule
    \end{tabular}
\end{minipage}
\par\vspace{10pt}
\noindent\begin{minipage}{\textwidth}
\centering
\includegraphics[width=\textwidth]{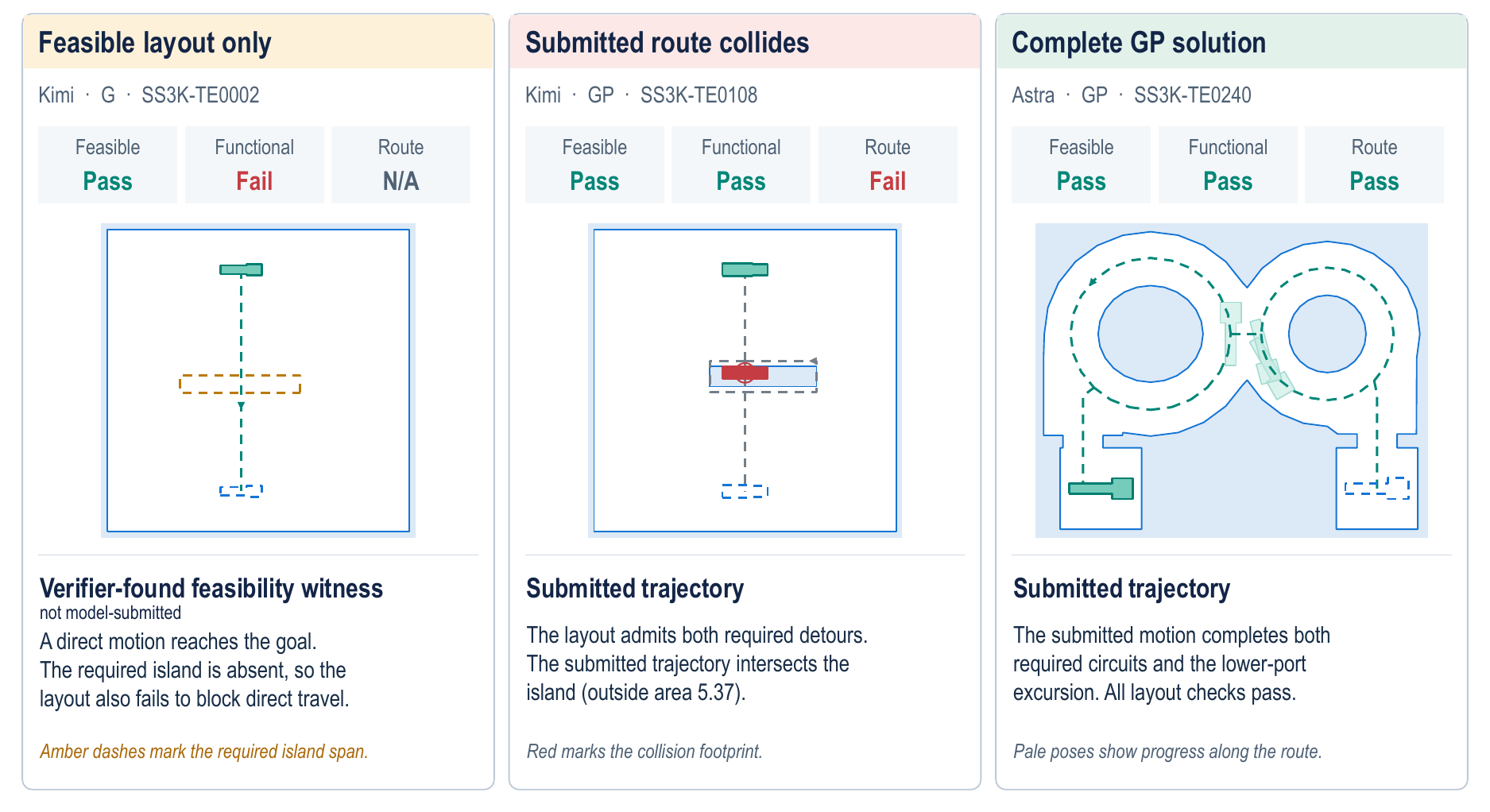}
\captionof{figure}{\textbf{Verifying construction and execution.} For G, the verifier finds a feasibility witness in the generated layout. GP also supplies a trajectory, which is checked separately.}
\label{fig:app_construction_examples}
\end{minipage}
\clearpage
\section{SofaSpace-Hard}
\label{app:hard}

\subsection{Task Types}

SofaSpace-Hard contains 534 tasks separate from the aligned 300-scene SofaSpace
test split: 473 planning tasks, 42 trajectory-validity judgments, and 19
feasibility judgments.  Table~\ref{tab:app_hard_tasks} reports the strict task-type
breakdown for all eight configurations.  Length-terminated outputs remain in the
scheduled denominator and contribute no strict success.

Of the 19 feasibility instances, four are feasible and fifteen are infeasible.
Each feasible instance has a verified complete motion.
Infeasibility is established by a case-specific geometric or timing certificate.
For example, one certificate separates the start and goal even in an outer approximation of collision-free configuration space.
Another shows that the body cannot fit in a region it must occupy at a specified time.

\par\medskip\noindent\begin{minipage}{\textwidth}
    \centering
    \captionof{table}{\textbf{SofaSpace-Hard strict accuracy by task type.}
    Counts are successes over the fixed task-type population.}
    \label{tab:app_hard_tasks}
    \small
    \setlength{\tabcolsep}{5pt}
    \begin{tabular}{lrrr}
        \toprule
        Model & Planning (473) & Trajectory validity (42) & Feasibility (19) \\
        \midrule
        Astra & 439/473 (92.81\%) & 40/42 (95.24\%) & 5/19 (26.32\%) \\
        Claude & 213/473 (45.03\%) & 39/42 (92.86\%) & 0/19 (0.00\%) \\
        DeepSeek & 9/473 (1.90\%) & 33/42 (78.57\%) & 0/19 (0.00\%) \\
        Gemini & 46/473 (9.73\%) & 29/42 (69.05\%) & 0/19 (0.00\%) \\
        GLM-5.3 & 4/473 (0.85\%) & 27/42 (64.29\%) & 0/19 (0.00\%) \\
        Grok & 10/473 (2.11\%) & 28/42 (66.67\%) & 1/19 (5.26\%) \\
        Kimi & 3/473 (0.63\%) & 26/42 (61.90\%) & 0/19 (0.00\%) \\
        Qwen 9B & 0/473 (0.00\%) & 11/42 (26.19\%) & 0/19 (0.00\%) \\
        \bottomrule
    \end{tabular}
\end{minipage}\par\medskip

\subsection{Mechanism Composition}

Hard covers all seven recorded mechanisms.  Atomic membership is overlapping,
while exact mechanism combinations form a partition.  The overlapping atomic
counts are C01: 242, C02: 77, C03: 151, C04: 83, C05: 203, C06: 136, and
C07: 54.  Table~\ref{tab:app_hard_combinations} lists the exact composition
counts; 183 tasks are single-mechanism, 290 are two-way compositions, and 61
contain three mechanisms.

\par\medskip\noindent\begin{minipage}{\columnwidth}
    \centering
    \captionof{table}{\textbf{Exact mechanism composition of SofaSpace-Hard.}
    Rows are mutually exclusive and sum to 534.}
    \label{tab:app_hard_combinations}
    \small
    \setlength{\tabcolsep}{4pt}
    \begin{tabular}{lr}
        \toprule
        Exact composition & $N$ \\
        \midrule
        C01 & 32 \\
        C02 & 12 \\
        C03 & 35 \\
        C04 & 33 \\
        C05 & 40 \\
        C06 & 25 \\
        C07 & 6 \\
        \midrule
        C01+C03 & 43 \\
        C01+C05 & 61 \\
        C01+C06 & 59 \\
        C01+C07 & 13 \\
        C02+C04 & 21 \\
        C02+C05 & 17 \\
        C03+C05 & 57 \\
        C04+C06 & 2 \\
        C06+C07 & 17 \\
        \midrule
        C01+C03+C05 & 16 \\
        C01+C06+C07 & 18 \\
        C02+C04+C05 & 12 \\
        C02+C04+C06 & 15 \\
        \bottomrule
    \end{tabular}
\end{minipage}\par\medskip

Table~\ref{tab:app_hard_combinations} assigns every task to one exact composition.
Atomic tag counts include a composed task under each of its mechanisms.

\newpage

\section{Representative Task Gallery}
\label{app:representative_gallery}

We selected 42 representative tasks. Figures~\ref{fig:app_aligned_views} and~\ref{fig:app_hard_interfaces} illustrate a subset of these tasks. We will publicly release the complete SofaSpace and SofaSpace-Hard task sets and their evaluation code by the camera-ready deadline.

 We drew one test scene from each of the seven exact mechanism compositions in SofaSpace and included its four aligned views (28 tasks). Figure~\ref{fig:app_aligned_views} shows the C04, C01+C03, and C07 scenes. The S and R prompts share the scene image, while G and GP present the fixed bodies and endpoints on an undesigned field. Each design prompt specifies its own functional requirements.
\par\medskip\noindent\begin{minipage}{\textwidth}
\centering
\includegraphics[width=\textwidth]{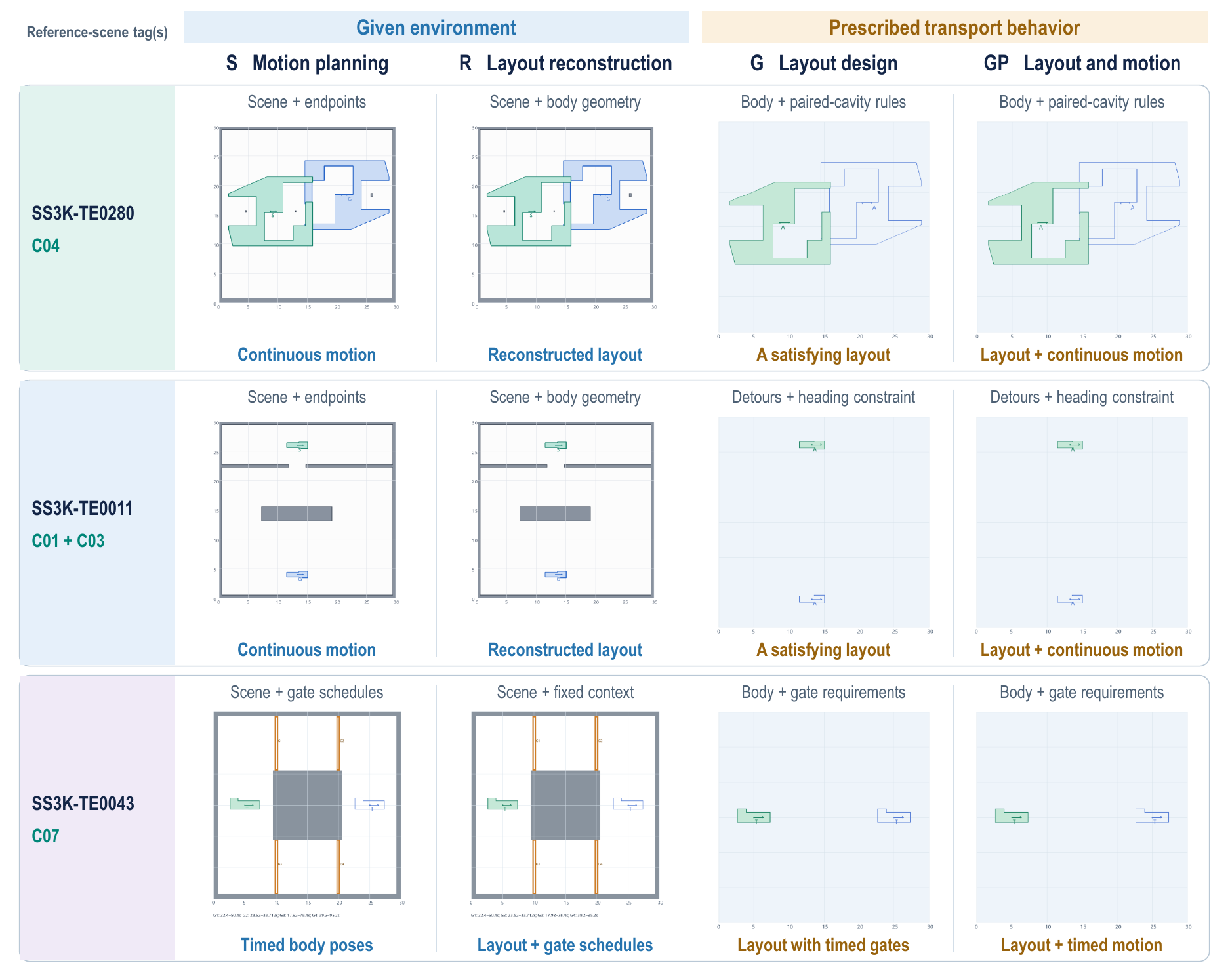}
\captionof{figure}{\textbf{Aligned task views.} Each row follows a single reference scene through S, R, G and GP. Its recorded mechanism tags describe the reference scene. The public G and GP prompts specify the functional requirements for the generated layout.}
\label{fig:app_aligned_views}
\end{minipage}\par

\newpage
The 14 Hard questions include the seven single-mechanism examples from Figure~\ref{fig:app_hard_examples}, three additional planning tasks with two or three mechanisms, and four additional judgment tasks. One S1 example is itself a trajectory-validity task. The additional judgment sample contains one positive and one negative instance for each of the two interfaces. Figure~\ref{fig:app_hard_interfaces} illustrates planning, trajectory validity, and feasibility using two questions per interface.
\par\medskip\noindent\begin{minipage}{\textwidth}
\centering
\includegraphics[width=\textwidth]{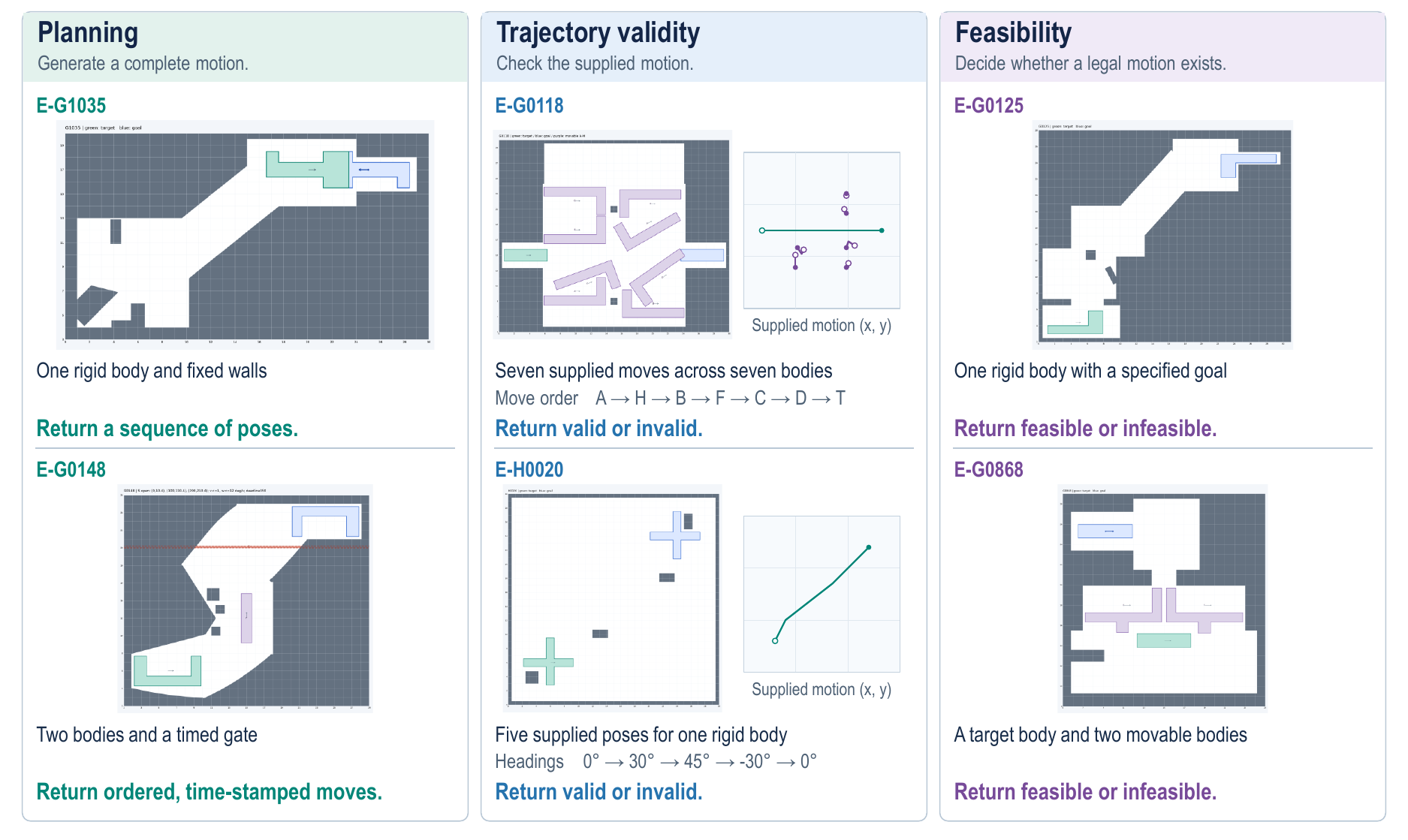}
\captionof{figure}{\textbf{Planning and judgment interfaces.} Each column shows two public Hard questions. The small trajectory plots reproduce the reference-point motions supplied for judgment.}
\label{fig:app_hard_interfaces}
\end{minipage}\par

\paragraph{Gallery sampling.}
For the remaining gallery examples, we sampled within the stated strata from sorted scene or task IDs using seed 2027, independently of model outcomes.

\end{document}